# RoSE: A Robotic Soft Esophagus for Endoprosthetic Stent Testing

Dipankar Bhattacharya[1,2] , Sherine Jesna V. A.[1,2], Leo K. Cheng[2,3,4], Weiliang Xu[1,2,4]

**Project page: https://bhattner143.github.io/rose-stent.github.io/**

**Abstract**

Soft robotic systems are well suited for the developing devices for biomedical applications. A bio-mimicking robotic soft esophagus (RoSE) is developed as an in vitro testing device of endoprosthetic stents for dysphagia management. Endoprosthetic stent placement is an immediate and cost-effective therapy for dysphagia caused by malignant esophageal strictures from esophageal cancer. However, later stage complications, like stent migration could weaken the swallow efficacy in the esophagus. The stent radial force (RF) on the esophageal wall is pivotal in avoiding stent migration. Due to limited randomized controlled trials in patients, the stent design and stenting guidelines are still unconstructive. To address the knowledge deficit, we have investigated the capabilities of the RoSE by implanting two stents (stent A and B) of different radial stiffness characteristics, to measure the stent RF and its effect on the stent migration. Also, endoscopic manometry on the RoSE under peristalsis has been performed to study the impact of stenting and stent dysfunctionality on the intra-bolus pressure signatures (IBPSs) in the RoSE, and further its effects on the swallowing efficacy. Each implanted stent in the RoSE underwent a set of experiments with various test variables (peristalsis velocity and wavelength, and bolus concentrations). In this paper, the conducted tests are representatives to show the application of RoSE to perform a wide-ranging assessment of the stent behavior. The usability of RoSE has been discussed by comparing the results of stent A and B, for various combinations of the test variables mentioned above. The results have demonstrated that the stiffer stent B has a higher RF, whereas, stent A maintained its RF at a low profile due to its lesser stiffness. The results have also implicated that a high RF is necessary to minimize the stent migration under prolonged peristaltic contractions in the RoSE. For the manometry experiments, stent A slightly increased the IBPS, but the stiffer stent B significantly decreased the IBPS, especially for the higher concentration boluses. It was found that if a stiffer stent buckles, it can reduce the swallow efficacy, and cause recurrent dysphagia. Therefore, RoSE is an innovative soft robotic platform capable of testing various endoprosthetic stents, thereby offering a solution to many existing clinical challenges in the area of stent testing.

**Keywords:** stent migration; radial force; soft robotics; robotic esophagus; biomimetics

## Introduction

Swallowing is a complex but orderly physiological process transporting saliva or food from the mouth to the stomach. The swallowing physiology and anatomy are elucidated with the critical insights from the in vivo studies.[1] Any esophageal impairment compromises the efficiency of

[1] Department of Mechanical Engineering, The University of Auckland, Auckland 1010, New Zealand.
[2] Riddet Institute, Palmerston North 4442, New Zealand.
[3] Auckland Bioengineering Institute, The University of Auckland, Auckland 1010, New Zealand
[4] Medical Technologies Centre of Research Excellence, Auckland 1010, New Zealand.

swallowing, is known as dysphagia.[2] Severe pathologies, for instance, benign esophageal strictures from various injuries, esophageal cancer, and esophageal perforations, distressing lumen patency explicitly leading to dysphagia. Unaccompanied by comprehensive care, a vicious cycle materializes as malnutrition and dehydration aggravating the dysphagia itself, subsequently increasing morbidity and even mortality.[3] The standard clinical practices involve evaluation of the etiology of the swallowing deficit using perception studies, and in vivo analysis, followed by intervention strategies for nutritional support.[4] Depending upon the severity, the method of maintaining nutrition, varies from oral dietary supplements, texture modified foods, implanting nasogastric feeding tubes, surgical correction, endoscopic dilation of the sphincters, and esophageal stenting.[2, 5]

The textured food modification is observed to maintain nutrition in the dysphagia and to improve swallowing safety and efficacy.[2, 5] In altered food swallowing, videofluoroscopy and manometry are considered as gold standards to study the temporal-spatial aspects of bolus geometry and intrabolus pressure, respectively. In an esophageal peristaltic transport of food bolus, the manometry recordings can be distributed into two intraluminal pressure segments. In the first segment, within the bolus fluid, the recorded pressure is solely due to the bolus hydrodynamic pressure, which is also known as the intrabolus pressure signature (IBPS). At the tip of the bolus tail, the pressure undergoes a transition from IBPS to esophageal direct contact pressure with the manometry catheter, which can be regarded as the second segment. The maximum IBPS occurs at the bolus tail tip, and after which, no bolus fluid exists, leading to direct contact pressure. Since pressure cannot be transmitted axially in the absence of bolus fluid; thus, these two segments are independent of each other.[6] Videofluoroscopy, and manometry techniques are used hand in hand to determine the maximum IBPS. The shortcomings of these techniques are X-ray radiation exposure and catheterization (inserting a manometry catheter) that accumulate to the efforts of swallowing and the general health of subjected individuals.[7]

Malignant and benign esophageal strictures from esophageal cancer, can be addressed with the endoprosthetic stent placement, commonly known as esophageal stenting.[8] The earlier use of uncovered stents for palliation was limited and contributed to *in situ* erosion, occlusion, and fistulation. However, with the advent of fully coated removable self-expandable plastic stents (SEPSs), self-expandable metallic stents (SEMSs), and biodegradable stents, remarked new applications with changeable success.[9] These stents designate a novel, alternative, immediate, and cost-effective therapy for managing adequate oral nutrition during dysphagia. A silicone covered SEMS is a tubular braided mesh of interwoven helical springs made of corrosion-resistant materials, like nitinol, stainless steel, and polymers.[10] These stents are proven to be efficient endoprosthetic management for both malignant and benign esophageal strictures, as they could hold open the esophagus and hence, relieve the impediments of compromised lumen patency in such cases (Fig. 1). Furthermore, the insertion procedure of a SEMS is less traumatic because of its flexibility, and it can be readily compressed into a smaller delivery system.[10-12].

However, this method of palliative treatment does have its inadequacies, and one of the significant shortcomings is stent migration, caused due to the interactions of an implanted stent with the continuous peristaltic contractile forces of the esophageal wall.[13] The stent deployment has reported about 30-50% long term success rates, with migration being the main associated complication in most of the failure cases.[14, 15] The efforts to mitigate pain and migration, to improve stent removability and flexibility, and to ensure stent patency have revolutionized the SEMS designs. The evolution of the stent design has made stenting a more effective treatment in both benign and malignant esophageal diseases. However, in the event of esophageal stent implantation, manometry and videofluoroscopy on the patients are less preferred for a compelling study on bolus geometry, IBPS, and stent inadequacies (such as stent migration), because the techniques are not very comforting for the patients, and there are major ethical concerns associated with such kind of studies. Despite recent technological innovations in this field, evidence to show which stent design is better than the other is confined to a few randomized controlled trials in patients with malignant esophageal strictures.[16]

The SEMSs have been found considerably similar to the esophageal prostheses.[17]According to the current US FDA review guidance for esophageal and tracheal prostheses devices bench testing of esophageal prostheses is needed to establish substantial equivalence.[18] The review guidance has recommended six different stent testing protocols. They are as follows: 1) Compression force testing: It measures the force required by the stents for compression. 2) Expansion force testing: It measures the force exerted by the stents during expansion. 3) Corrosion testing: It tests the compatibility of the stent materials with the corrosive environment in the esophagus. 4) Tensile strength tests: If a stent includes a bonded or welded part, then this test needs to be performed. 5) Deployment testing: It validates the accuracy and repeatability of the stent delivery system. 6) Dimensional testing: It verifies the dimensional reproducibility of the stents after deployment. The first four tests are related to stent performance after its deployment, and the last two tests are relevant in terms of the insertion procedure. The compression and expansion force testing of the stent includes stent radial force (RF) measurement, which is crucial in designing a stent for maintaining the *in situ* lumen patency, which is, unfortunately, still an unknown parameter due to the poorly understood association of the RF and clinical outcomes (Fig. 1A).[11, 12] Since the adequate data on migration issues, critical complications, and morbidity in stent deployment, as well as extraction, are lacking, the stenting guidelines need clarity and further research. After the stent deployment, the internal pressure acting on the esophageal wall is the opening RF applied by the stent per unit contact area of the stent ($\pi d_e l_e$). The internal pressure accounts for the stress in the esophageal wall, known as the circumferential (hoop) stress ($\sigma_\theta$, Fig. 1B).[19] RF assures proper fixation of the stent in the esophagus, and an essential factor in limiting stent migration.

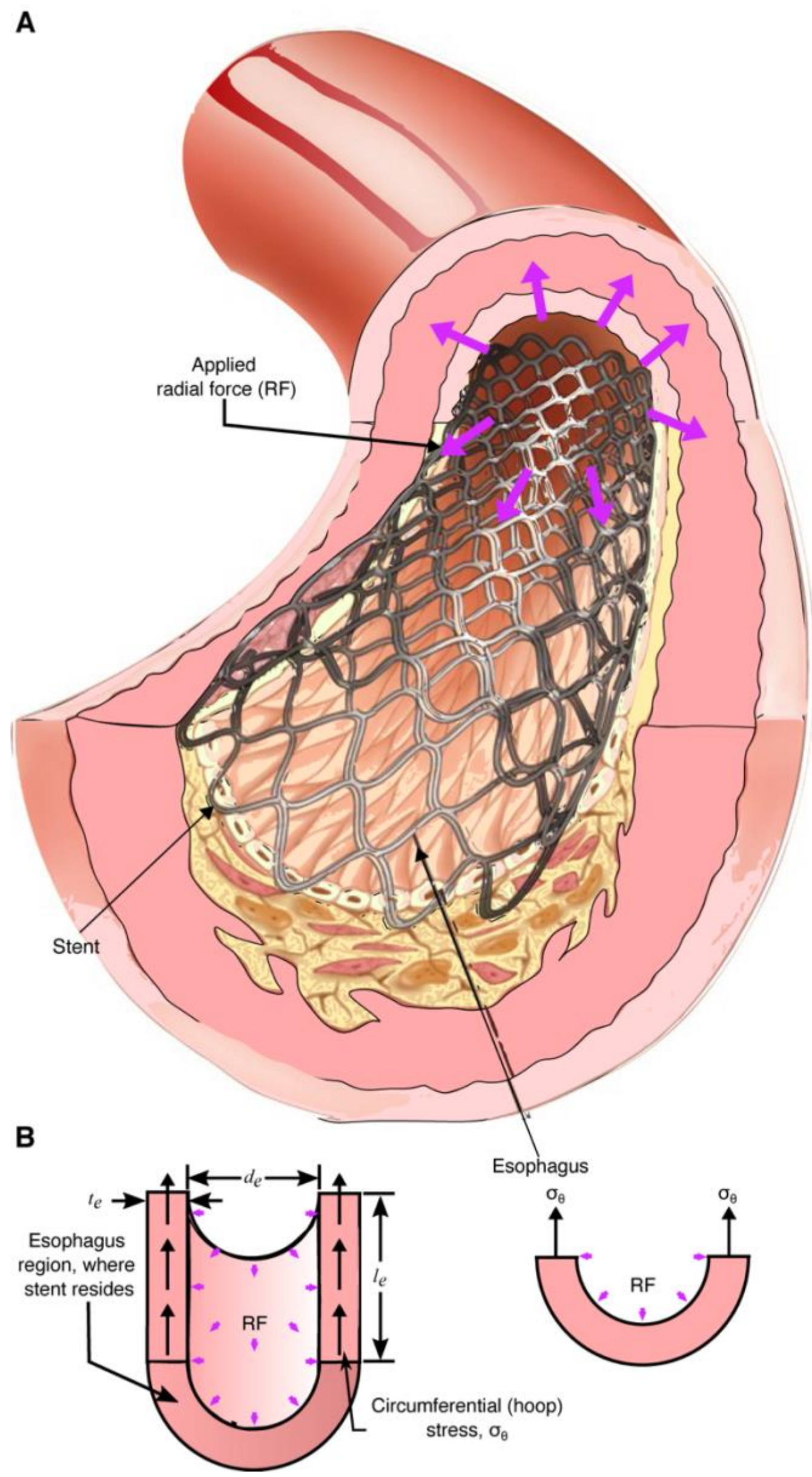

**Fig. 1. Cross-sectional view and free body diagram of an esophagus with an implanted stent**. (A) Schematic of isometric view showing the cross-section of an esophagus with an implanted stent. The purple blue arrows represent the radial force (RF) applied by the stent. (B) Free body diagram of the esophagus region, where the stent resides. The hoop stress, $\sigma_\theta$ is given by $pd_e/2t_e$, where $p$ is the applied pressure, and equal to $\mathrm{RF}/\pi d_e l_e$.[19]

While a considerable amount of literature is available on novel stent designs, little attention has been paid on comparing the differences in RF and its clinical impacts among the available stents in the market. An investigation in such a direction could potentially play a pivotal role in dealing with the migration and symptoms of pain. The mathematical modeling of the stent interaction with the vessel wall, provided in various studies, could be useful for predicting the mechanical properties of the stent. Still, the studies fail to embrace the esophageal peristalsis complexity and precisely replicate the association between the esophageal wall and the stent. The analytical model of the stent based on the theory of virtual work by Jedwab and Clerc [20] is useful for predicting the mechanical properties of a stent (such as the RF), but it model fails to provide any implication of stent migration with its mechanical properties. Hirdes et al.[11] evaluated RF of several commercially available stents, but the author could not offer clear evidence to show that high stent RF can prevent stent migration. Likewise, numerous numerical studies have focussed solely on explaining the mechanical deformation of the stent inside the host tissue.[21, 22] Very few numerical models have examined the interaction between the stent and the esophagus under peristalsis.[15, 23] Garbey et al.[15] analyzed stent migration in a numerical esophageal model with stent flare design, stent length, diameter, and radial stiffness but not directly with stent RF. Mozafari et al.[23] identified flared stent ends, and a higher frictional coefficient of the stent can have a considerable impact on mitigating stent migration. The study was done with an FEA built esophagus, under axial pulling (peristalsis was not considered). In this study, from henceforth, stent refers to covered SEMS.

Soft robotics in vitro models has proved to be a complementary and supplemental approach to mathematical models, and clinical studies to investigate the human physiology and to validate medical procedures.[24] The excellent adaptability of soft robots to their environment has been explored in the recent decades, leading to the development of next-generation soft materials and soft actuators, soft stretchable electronics, and control, and processing.[25-28] Soft robotics also widens the opportunities for developing devices in the field of biomedical applications, such as therapeutic devices,[29] rehabilitation devices, and prostheses,[30] devices for training and biomechanical studies.[31] With Kobayashi et al.[32] pioneering the robotic swallowing simulator works, the studies in silico and in vitro model of esophageal phase have provided reasonable mechanistic insights for many interesting archetypes observed in the in vivo investigations of the human swallow physiology.[33] Hence a bio-mimicking robotic soft esophagus (RoSE) is developed as an esophagus simulator with extended biomedical applications.[34]

In the medical domain, testing bolus formulation and transport are hindered by the inter-person swallow and the inter-swallow variability in human test subjects.[35] Variations in muscle actuation strength and peristalsis wave speed greatly influence the manometry captured IBPS in the subjects; thus, testing and measuring the swallow efficacy in a man is qualitative. In the mathematical field, the modeled boluses do not reflect the complicated behavior of bolus materials such as multi-phase flow and non-Newtonian property.[36] Besides, it is challenging to model the shear fields generated by peristaltic actuation and their associated time-shear dependent behavior. The soft-bodied robot, RoSE (Fig. 2) can physically mimic the human swallowing action, by generating peristaltic waves (Fig. 2), to transport the food bolus along the conduit. Unlike other in vivo methods, RoSE offers a more steady swallowing behavior and does not hold the risk associated with testing actual human subjects.[34] RoSE bridges the gap between clinical and mathematical modeling fields, and it is proposed to deform materials to achieve clinically significant rheometry more faithfully. Besides, RoSE could vary the bolus parameters and the peristaltic parameters (such as wave speed and wavefront length) independently, which is crucial to comprehend the effect of these variations as a fit to a broader set of similar pathological paradigms. To mimic the human swallowing behavior, synthetic bolus formulations were tested on the RoSE with manometry and videofluoroscopy. The manometric pressure profiles achieved are comparable to human swallowing behavior.[37] Since RoSE can physically mimic the human swallowing behavior; thus, instead of actual patients, RoSE can be used to conduct the study on the various stent designs and their inadequacies before implanting them in patients with malignant and benign esophageal strictures. Besides, to evaluate swallow efficacy, RoSE can also be used to study the effect of stenting on IBPS of different texture modified foods for dysphagia patients.

This study aims to investigate the following capabilities of RoSE, 1) to perform experiments on two stents having different radial stiffness characteristics for evaluating their respective RFs,

2) to study the effect of the RFs on the migration of the stents, 3) to learn the impact of the reduction of RoSE conduit diameter caused by stent deployment on IBPS, and 4) to analyze the effect of stent dysfunctionality on bolus transport, and hence, swallow efficacy.

The RoSE stent testing protocols provides a novel in vitro platform to perform a wide-ranging assessment of the stent behavior. Besides, the protocols are capable of generating spatio-temporal peristaltic waves of different characteristics in the RoSE conduit. In this study, commercially designed stents are subsequently deployed in the RoSE and analyzed to validate the above-vested capabilities of the Mechatronics device. Along with the circumferential loading of the stent, studies on the effect of stent implantation on bolus transport, under peristaltic contractile forces are done by controlling the RoSE conduit deformation.

The analysis of the bolus pressure signatures with stenting in a soft-robotic *in-vitro* platform, like the RoSE, has extended our knowledge of swallow efficacy after stent implantation during dysphagia. Besides, this study has also done endoscopic manometry in the presence of stents, and it has also considered esophageal peristalsis for stent-related measurements. To the best of our knowledge, studies correlating the effect of stent implantation and dysfunctionality on swallow efficacy with parameters (such as IBPS and IBPS gradient) that can be measured, still does not exist. Little is known about the IBPS and IBPS gradient due to the implantation of stents and their dysfunctionality. Much of the research up to now are clinical studies. Investigating stent RF has been a continuing concern within the clinicians and the scientific community. It has been challenging to fully predict the clinical outcome of a stent based on its RF and bolus pressure signature data.[11] However, the presented results show that RoSE can provide a platform to assess various stent behaviors, that can help the researchers and the endoscopists to elucidate the patency of the stents in the occurrence of unfavorable events, during and after the stent implantation. Also, the results comparing the efficiency of the stents could aid the endoscopists in the selection of a suitable stent for an individual patient, which is otherwise guided solely by their experience and availability of the stents. The results will also aid in the optimization of future stent designs to improve the behavior of the stent during deployment and to mitigate migration.

## Materials and Methods

### *RoSE actuator, design and materials*

RoSE actuator has 12 layers ($L_1, L_2, \ldots, L_{12}$) of regular and repeating pneumatic hollow chambers arranged horizontally along the axis of the device, with a whorl width of 10 mm each (Fig. 2). Each layer has four chambers arranged axis symmetrically, and have whorl boundaries of 5 mm thickness with theoretically identical properties (Fig. 2). When pressurized with air, the four chambers expand and closes the cross-sectional area of the food passage and push off the housing structure to cause the radially occlusive peristaltic motion, mimicking circular muscle activation. Though the actuation in RoSE is distributed, by reducing the actuator chamber width, whorl boundary and conduit thickness, continuous peristaltic actuation is achieved (Fig. 2). The actuator design specifications are based on the FEA analysis conducted by Chen et al.[38] Table 1 provides the quantitative and qualitative characteristics of RoSE, which are required for performing stent RF and migration, and stent implanted RoSE swallow efficacy testing.

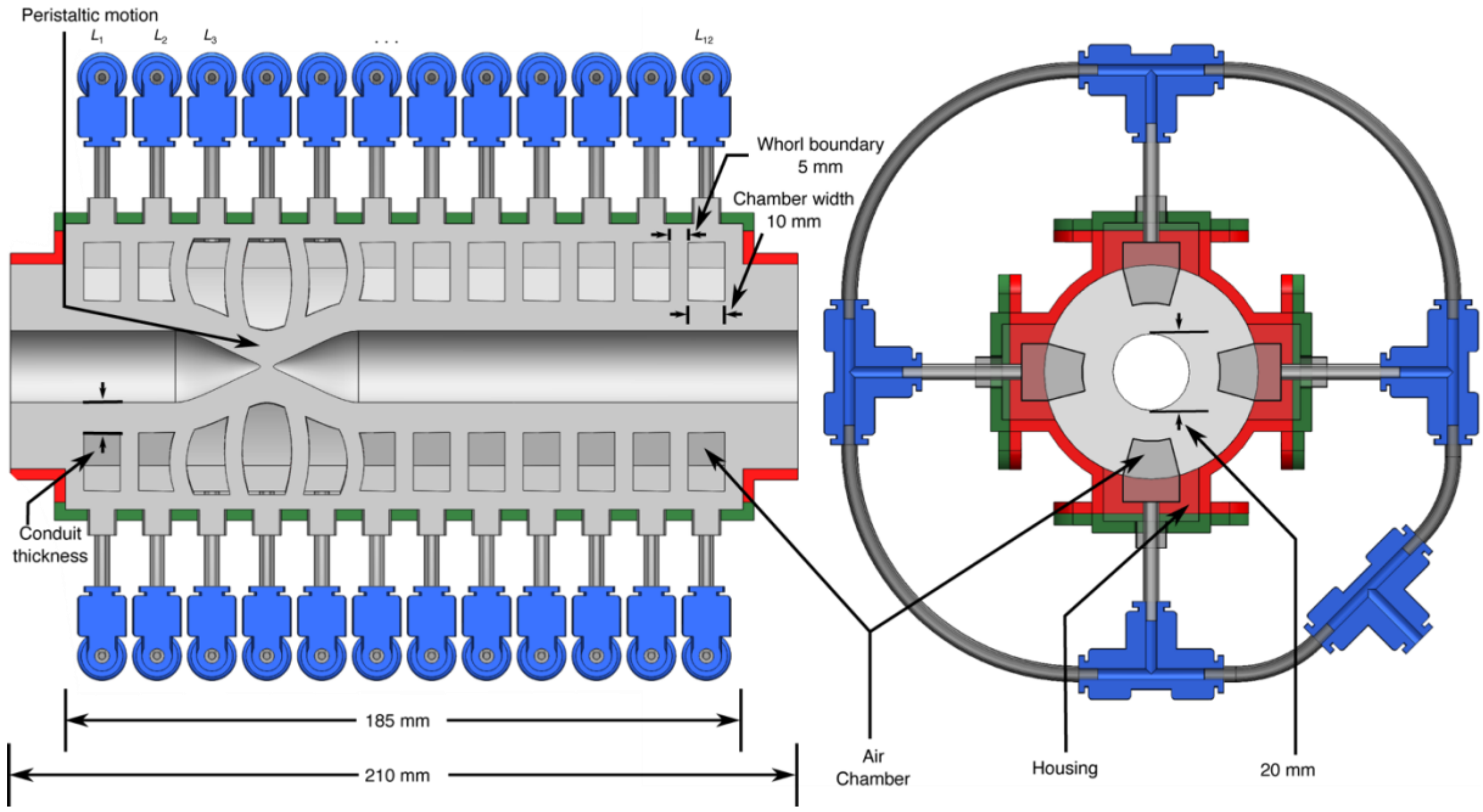

**Fig. 2. Side and top sectional view of robotic soft esophagus (RoSE).** The left and right drawings reflect the longitudinal and axis-symmetric arrangement of the chambers, respectively.

**Table 1: Quantitative and qualitative characteristics of RoSE**

| **Quantitative** | |
|---|---|
| **Attributes of RoSE** | **Magnitude** |
| Conduit length | 210 mm |
| Actuating conduit length | 185 mm |
| Conduit diameter at rest | 20 mm |
| Conduit diameter range | 0 – 20 mm |
| Applied pneumatic pressure | 500 KPa |
| Number of axial levels (layers) | 12 |
| Chambers per layer | 4 |
| Operating pressure range | 0 – 71.5 KPa |
| Minimum peak occlusion pressure (peak contact pressure) | 16 KPa |
| Pneumatic pressure required to achieve medical wave seal pressure (15 KPa) | 64 KPa [34] |
| Wave velocity | 20, 30, and 40 mms$^{-1}$ |
| Wavefront length | 40, 50, and 60 mm |
| Data acquisition resolution | 8 bits |
| Frequency range | 400 kHz – 1.7 MHz |
| Migration measurement resolution | 2 mm |
| Migration measurement range | 0 – 100 mm |
| | |
| **Qualitative** | |
| Actuation type | Symmetric, peristaltic |
| Bolus transport type | Peristaltic |
| Device compliance | Compliant, continuous |
| Muscle activation | Overlapping-sequential |

RoSE is fabricated using custom-designed mold and housings, which are rapid prototyped via fused deposition of acrylonitrile butadiene styrene (ABS) plastic (Fig. 3A). An RTV silicone rubber material (Ecoflex 0030, Smooth-on, USA) is chosen to build the actuator (Fig. 3B). The silicone rubber was chosen for its low 100 % modulus and pour viscosity, high tear resistance, and surplus elongation at break (900 %), which is needed during actuator deformation. In addition, RoSE design is based on the idea of actuating many layers in a subsequent overlapping manner, which mimics the esophagus in recruiting muscles in a rostro-caudal way, to achieve peristaltic transport smoothly. The multi-casting fabrication procedure of the RoSE is shown in Fig. 3A to F.

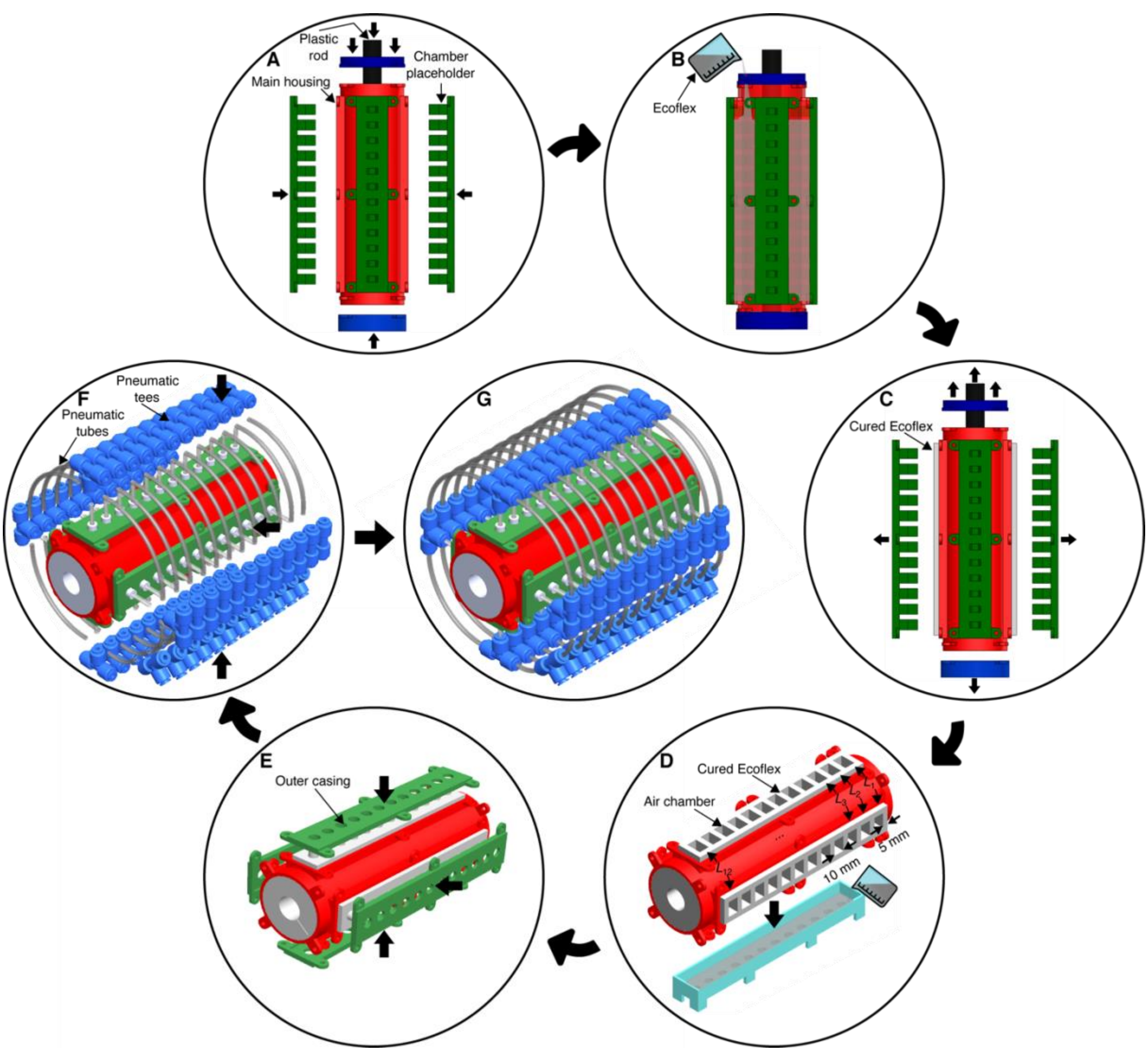

**Fig. 3. Multi-casting fabrication procedure of RoSE.** (A) The actuator mold and housing are rapidly prototyped by the printing method from ABS plastic (Elite Printer, Dimension, USA), and the initial assembly is built. (B) In vertical orientation, the conduit and the chambers are constructed by casting silicone rubber (Ecoflex 00-30, Smooth-On, USA). (C) After the silicone cured, the chamber place holders and the plastic rod is removed. (D) In horizontal orientation, four outer castings of silicone are made to seal the chambers from the sides. (E) Four outer casings are added to confine the chamber deformation within the conduit. (F) Pneumatic tubes and tees are connected to provide air pressure to the whorl of chambers (G) Final assembly of RoSE.

*RoSE firmware protocol*

Custom firmware modules, written in Python 3.7, are developed on Raspberry Pi 3B+ to assert the robot, with independent, continuously variable, pressure input (Fig. 4). The RoSE firmware module is divided into two significant sub-modules: A) Symmetric actuation protocol, for simultaneous inflation and deflation of the RoSE conduit, to continuously record the radial force during contraction and expansion of stents. B) Peristaltic actuation protocol, for generating spatio-temporal waves in the RoSE conduit, to learn the stent migration and effect of stenting on the IBPS (Fig. 4). The user can select an appropriate testing protocol module by following the prompts on the IDLE (Python's integrated development and learning environment) shell window. Each testing protocol (sub-module) selects a CSV file where 8-bit digital values corresponding to a pressure range are stored for $L_1, L_2, ..., L_{12}$ layers of RoSE.

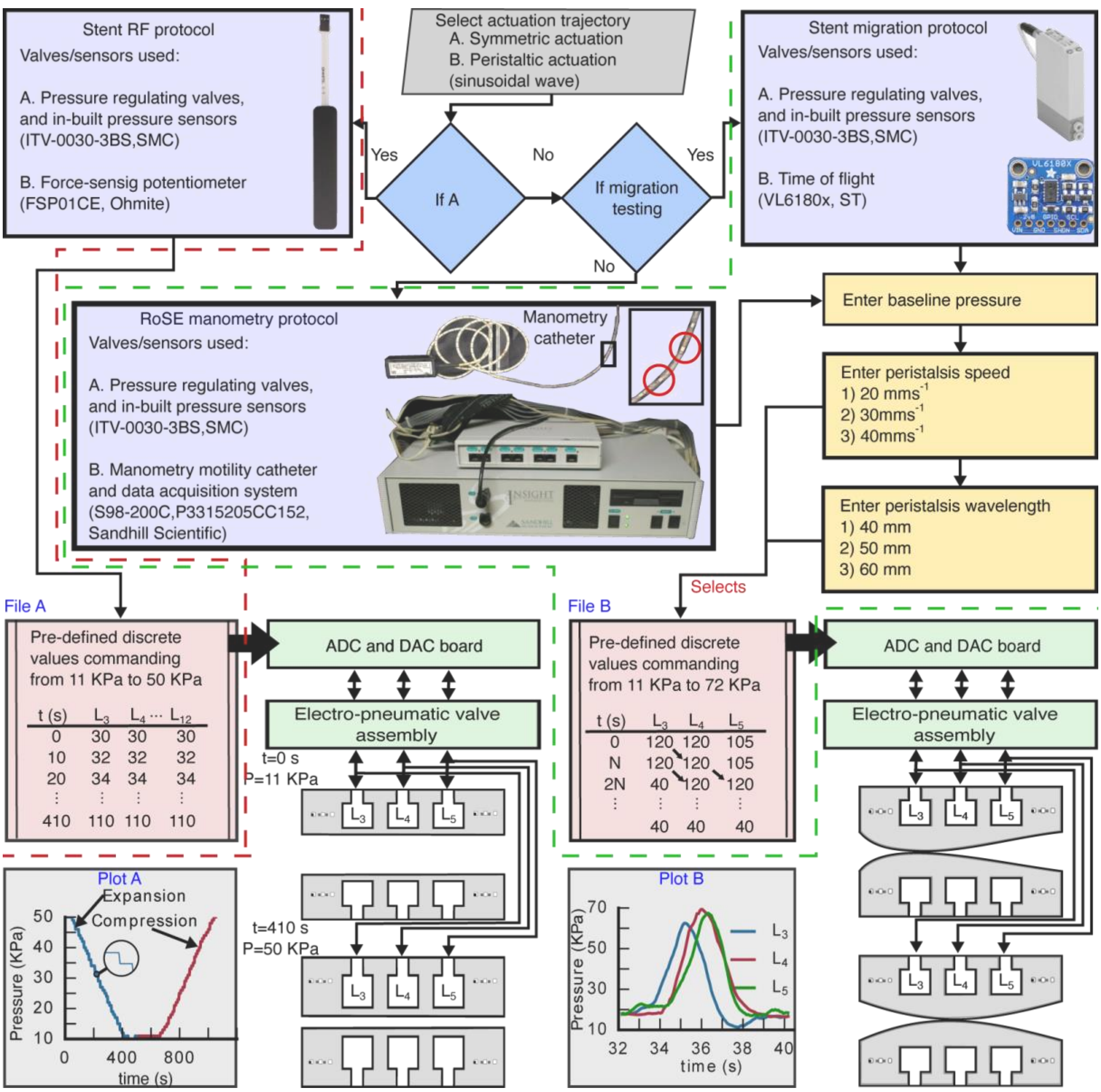

**Fig. 4. Flow chart illustrating system description of RoSE.** RoSE firmware protocol is capable of generating two different actuation trajectories (symmetric and peristaltic actuation) based on the input commands of the user. The firmware protocol for symmetric and peristaltic actuation, developed on Raspberry Pi 3B+, is enclosed within the red and green dashed lines, respectively. The blocks outside the dashed lines represent the RoSE hardware, which consists of interfacing ADC and DAC board, assembly of pneumatic valves, and sensors such as TOF and valve in-built pressure sensor.

Pneumatic pressure to RoSE is regulated in open-loop by an electro-pneumatic interface which consists of a series of 12 pressure regulating proportional valves (ITV-0030-3BS, SMC, Noblesville, IN, USA), and an interfacing ADC (MAX11605, Maxim Integrated, San Jose, CA,

U.S.) and DAC (AD8802, Analog Devices, Norwood, MA, USA) board, interfaced with the Pi (Fig. 5). The Pi communicates over SPI and I2C with the DAC and ADC, respectively, to perform assertion and feedback measurement. The DAC converts 8-bit digital values, stored in the CSV files, to voltage levels, which controls the valves. The ADCs are then used to receive various sensor outputs such as in-built valve pressure sensor and force sensing potentiometer (FSP01CE, Ohmite, active area – 10 x 13 mm$^2$). The valve pressure sensors are used to record the chamber pressure-time trajectory (Plot A and B, Fig. 4).

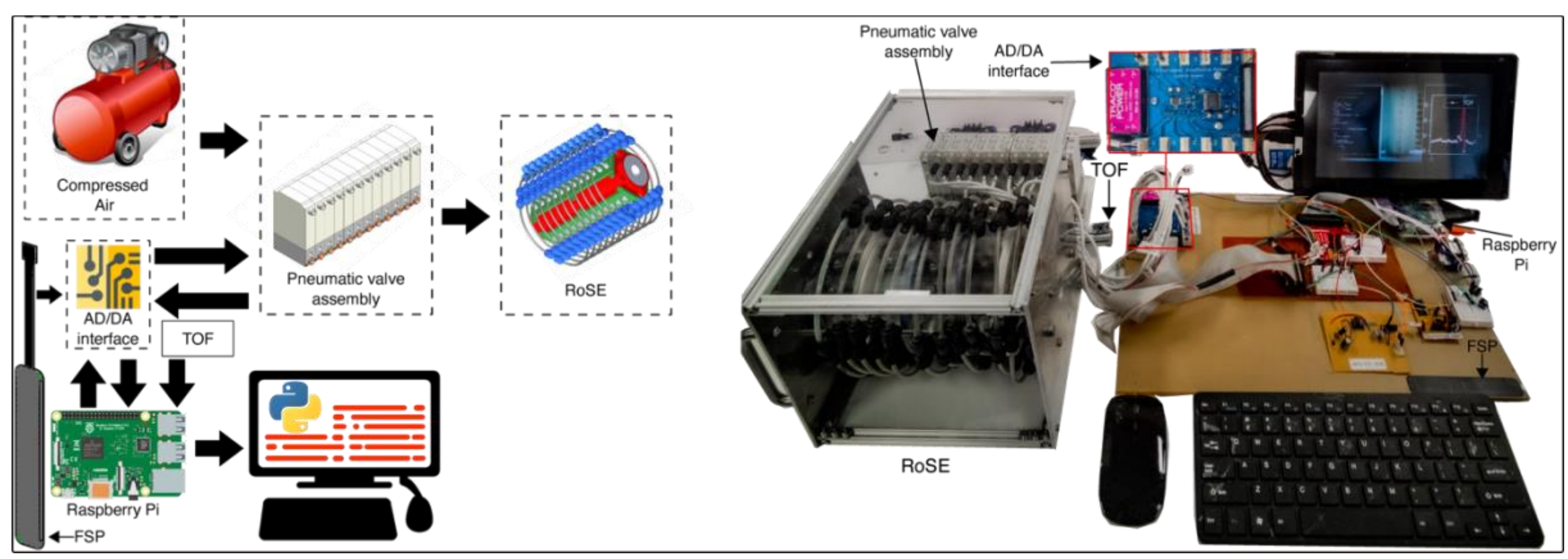

**Fig. 5. Schematic and image of the electro-pneumatic interface controlling RoSE.** An assembly of 12 proportional pneumatic valves, sourced by compressed air supply, are implemented to control the pressurization of 12 layers of RoSE. The valves assembly is controlled by an custom-built interfacing ADC and DAC board, connected to a Raspberry Pi 3B+. Firmware protocols are developed on the Pi with Python (v3.7) to actuate RoSE with independent, continuously variable pressures.

In symmetric actuation, layers $L_1$ to $L_{12}$ of the RoSE were actuated, with the same pressure input, from 11 KPa to 71.5 KPa in 1 KPa increments. In a CSV file, 8-bit digital values corresponding to the pressure range is stored (File A, Fig. 4). Each value was held for 5 s to check the repeatability of the sensors (data was collected at 1-second interval) and to allow the chambers to reach their settling deformation.

For manometry, and migration experiments, three adjacent RoSE layers are sinusoidally actuated with multi-level pressures to achieve the spatial-temporal aspects of peristalsis waves. In the peristaltic actuation protocol, the displacement of the conduit can take the shape of a sinusoid with complete occlusion at the point where the minima of the sinusoid occur. By displacing the sinusoid from $L_1$ to $L_{12}$ in small increments, continuous peristalsis wave in RoSE was achieved. The sinusoid can be represented by (1).

$$H(x,t) = \begin{cases} \epsilon, & x < ct \\ \epsilon + \frac{\alpha}{2}\left(1 - \cos\left(2\pi\frac{x-ct}{\lambda}\right)\right), & ct \le x \le ct + \frac{\lambda}{2} \\ \epsilon + \alpha, & ct + \frac{\lambda}{2} < x \end{cases} \quad (1)$$

where $H(x,t)$ is the time-dependent conduit radius (mm), ε is the minimum conduit radius (mm), α is peak-to-peak peristalsis wave amplitude (mm), $c$ is peristalsis wave velocity (mmps), λ/2 is sinusoidal wavefront length (mm), $x$ is conduit axial displacement (mm), and $t$ is time (s) .

From the IDLE shell window, the user can define the peristalsis wave speed ($c$) and wavefront length ($\lambda/2$). Each combination of the wave speed and wavefront length selects a specific CSV file (File B, Fig. 4) that stores a set of 8-bit digital values to impose a time-variant pressure trajectory between adjacent RoSE layers (Plot B, Fig. 4), so that RoSE conduit can displace in a peristaltic manner. For complete occlusion, an operating pressure range of 11 KPa to 71.5 KPa was used. The actuation and the continuity of peristalsis pattern are verified using medically inspired techniques such as videofluoroscopy,[37] articulography,[34] and image processing of a quarter RoSE.[39]

*Stent configuration*

Fifteen commercial, covered SEMSs having distinct structures, cover materials, and cover material patterns, and similar dimensions were initially tested for stent migration in RoSE. Based on the maximum and minimum recorded migration, candidate stents: stent A and B were selected for further measurement and analysis of different stent-related parameters (such as stent RF, radial stiffness and migration), and bolus pressure signatures in RoSE with and without the stents.

The commercial stent A and B with different radial stiffnesses, under test, are cylindrical with flares at both ends, covered in silicone, and braided from a single thread of highly elastic nitinol wire of 0.10 mm diameter (Fig. 6). When fully expanded, the main body of the stents measured 110 mm in length and 23 mm in diameter. The flare ends are 10 mm in length and 5 mm wider in diameter than the body.

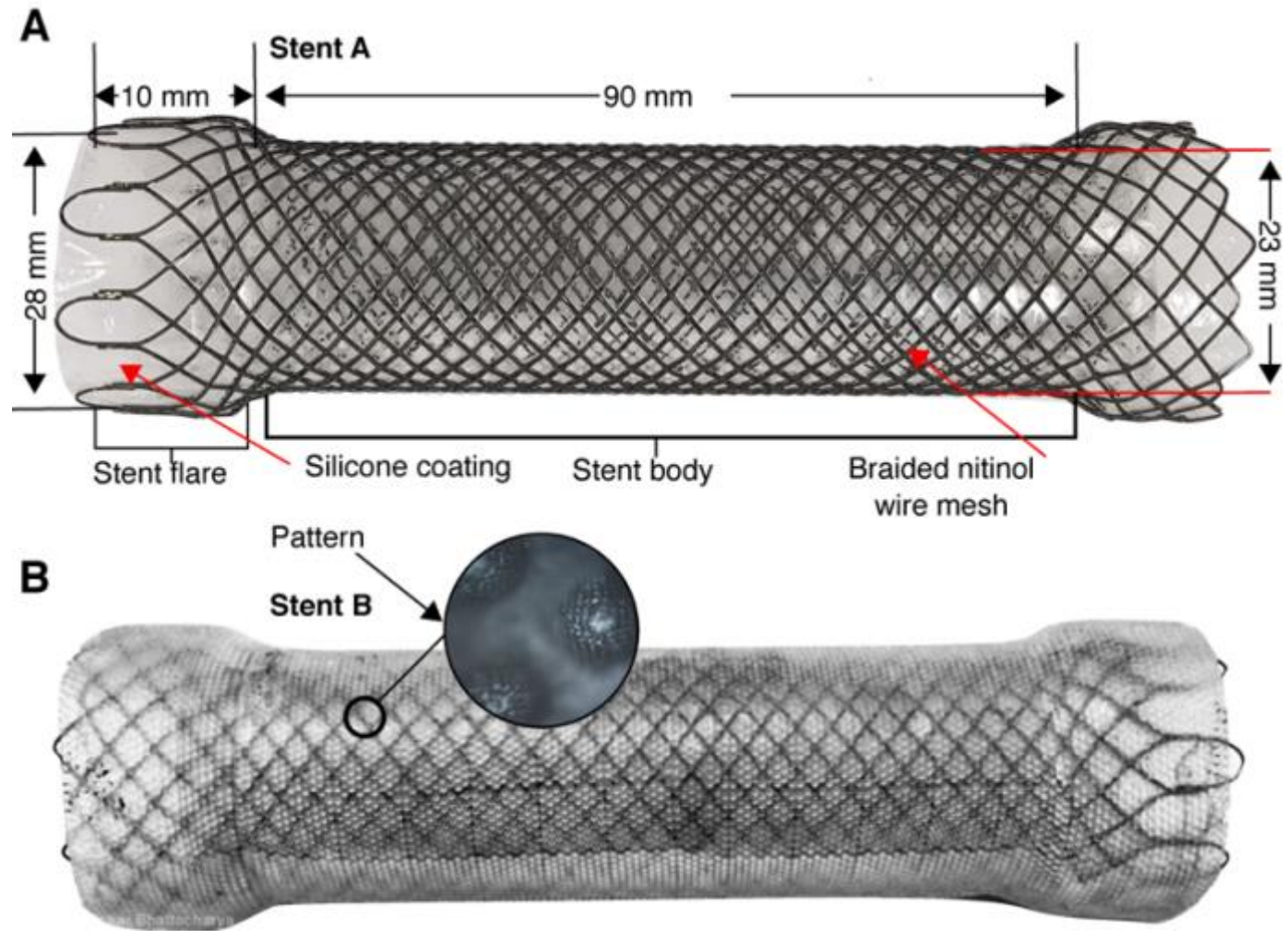

**Fig. 6. Silicone covered self-expandable metallic stents (SEMSs).** (A, and B) Stent A and B have a mean radial stiffness of $1.55 \pm 0.24$, $3.13 \pm 0.53$ Nmm$^{-1}$ respectively. Although both the stent has the same nitinol braided wire mesh configuration, the silicone coating thickness and pattern are different. Stent A has no pattern on its silicone cover, while stent B has much thicker silicone cover with a pattern on it.

*Food bolus*

Synthetic boluses of starch-thickened water (Altrix Rapid Thickener, Douglas Nutrition Ltd, New Zealand), have been used as a clinically significant substitute to masticated boluses throughout experimentation. Boluses were formulated on a concentration basis (72, 108, 144 $gL^{-1}$) by thorough mixing, after which they were left to stand for an hour to stabilize. The settling time allowed the starch granules to take up water and achieve a stable structure which directly affects the rheological properties of the boluses. These boluses formulations covered the range from syrup thin to pudding thick based on the product specification as per International Dysphagia Diet Standardization Initiative (IDDSI) specification.[5] All the formulations of the starch-thickened water have exhibited shear thinning behavior. Due to this non-Newtonian effect, the viscosity will be depended on the flow characteristics such as peristalsis wave velocity. In this study, the three used bolus concentrations mixes are labeled as bolus I, bolus II, and bolus III, respectively (Table 2).

**Table 2. Characteristics of the starch-thickened food boluses.**

| **Bolus** <br> **($gL^{-1}$)** | **Thickener Concentration** | **Viscosity (Pa.s)** |
|---|---|---|
| I | 72 | 0.62 |
| II | 108 | 1.08 |
| III | 144 | 1.55 |

*The protocol of stent RF and migration measurement*

To ensure prolonged functionality of the esophageal stent in the face of the repetitive peristaltic contractile forces associated with the peristalsis, a high RF and fatigue strength, corrosion resistance, and deformability are required. After the deployment of the stent in RoSE conduit (Fig. 7 A, to C), the stent apply an opening radial pressure to the conduit wall, resulting in a hoop (circumferential) stress in the wall, increasing its diameter.[19] A similar phenomenon occurs in a stent implanted esophagus (Fig. 1). The pressure is associated with the stent RF on the RoSE conduit wall as it tries to expand back to its initial diameter. The pressure can be defined as the stent opening RF acting on the conduit wall per unit cross-sectional area of the RoSE conduit (Fig. 7C) whereas, the hoop force (HF) is the hoop stress multiplied by the cross-sectional area of the RoSE conduit wall, where the stent resides (footprint of the stent). Similarly, radial pressure on the stent imposed by the RoSE contraction, generates RF and HF on the stent wall (Fig. 7 D to F).

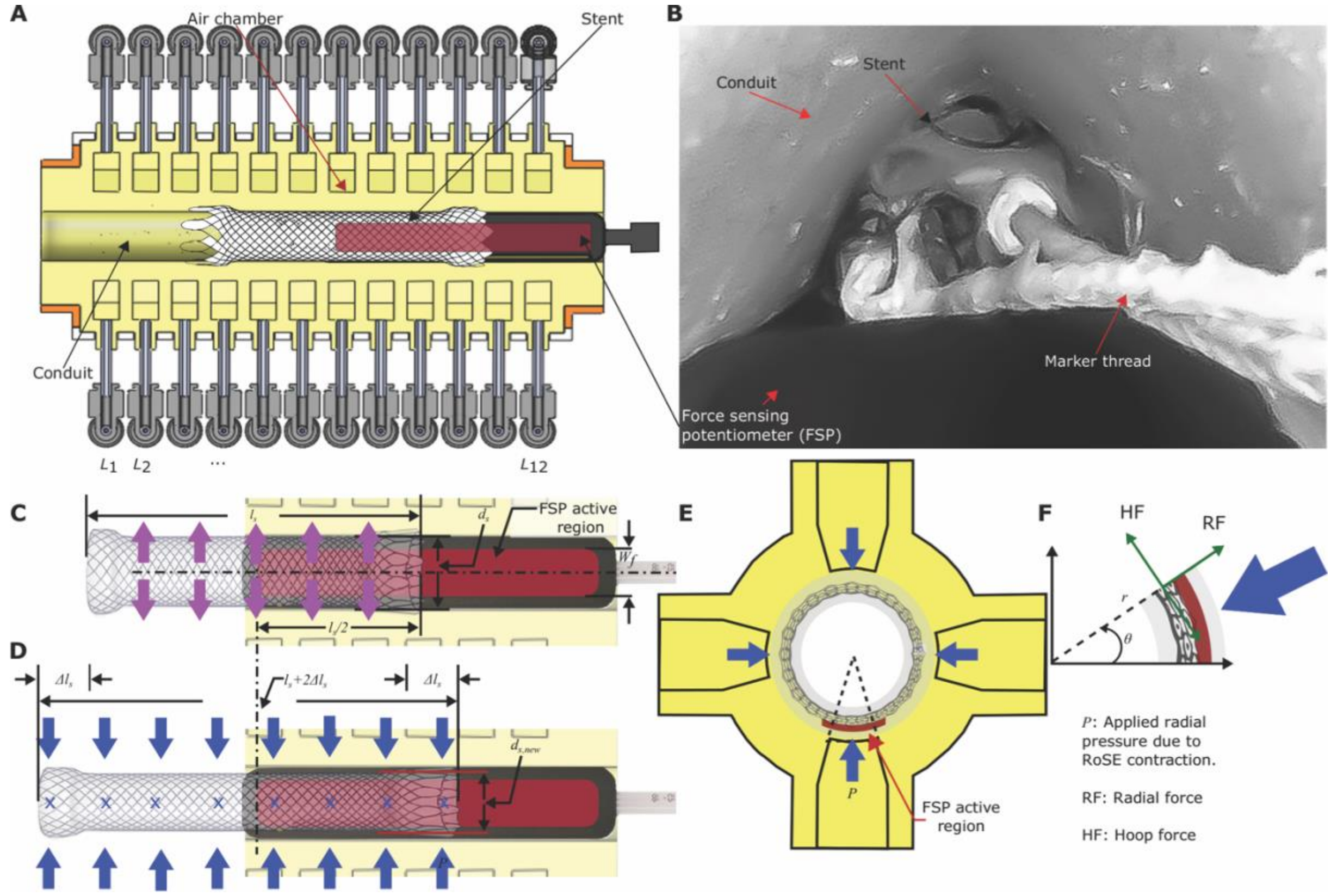

**Fig. 7. Experimental setup to measure the radial force (RF) exerted by a stent on RoSE conduit.** (A) Schematic of the experimental setup used for measuring the stent RF. Firstly, the testing protocol comprised of implanting a force-sensing potentiometer (FSP), followed by the subsequent deployment of the stents. Secondly, all the RoSE layers ($L_1, L_2, ..., L_{12}$) were inflated and deflated cyclically with the same air pressure to generate cylindrical contractions, and expansions respectively. (B) Endoscopic image showing the cross-sectional view of the RoSE conduit during contraction, captured from the distal end of the RoSE. (C) Schematic showing the position of the stent and the FSP inside the RoSE conduit. The purple arrows represents the opening force applied by the stent. (D, and E) Schematic showing the side and top view of the stent-FSP implanted RoSE under contraction (blue arrows are representing the loading on the stent). (F) Free body diagram of the stent to represent RF and HF.

The RF can be classified into: 1) Radial resistance force (RRF): RRF can be defined as the stent resistance to the loading when the RoSE conduit (esophagus) is contracting (RRF is the RF shown in Fig. 7 D to F). 2) Chronic outward force (COF): When the RoSE conduit (esophagus) is expanding, the force exerted by the stent to regain its original diameter during its unloading (Fig. 7C) is known as COF. RRF and COF are forces associated with the loading and unloading of the stents, respectively.

By using a force-sensing potentiometer (FSP01CE, Ohmite, active area – 10 x 13 mm$^2$) sensor, the RF exerted by the stent on the RoSE conduit wall was measured (Fig. 7). After lubricating the conduit of the RoSE with artificial saliva (Aquae Dry Mouth Spray, Hamilton), the stent under the test was deployed from the distal side of the RoSE by using an intruder sheath and a retrieval thread fastened to the proximal end of the stent. While holding the stent with a rat tooth forceps from the distal side of the RoSE, and by withdrawing the intruder sheath gradually from the proximal end of the RoSE, the stent was released slowly in the conduit. Owing to the stent stored strain energy, it self-expanded and exerted an RF on the conduit wall, and hence on the FSP sensor, which readily hooped around the stent (Fig. 7E, Movie S1). To achieve uniform axial displacement during compression, both the stent ends were allowed to move freely.

To ensure a consistent reference configuration, the stent center was positioned to coincide with the distal end of FSP active region having a width of $W_f$ (Fig. 7C). The 3 mm smaller diameter of the RoSE conduit from the stent body caused transverse loading of the stent which deformed the stent both axially and radially. If $l_s$ and $d_s$ are the new deformed length and diameter of the stent respectively, and if $F_f$ is the force recorded by the FSP, then stent radial pressure ($P_s$) can be given as:

$$P_s = \frac{F_f}{\left(\frac{l_s}{2}\right) W_f} \tag{1}$$

By considering uniform diameter throughout the stent, the lateral surface area of the stent is given by, $A_s = \pi d_s l_s$. Hence, the stent RF ($F_s$) can be written as:

$$F_s = 2\pi \frac{F_f}{W_f} d_s \tag{2}$$

Eq. (2) is also applicable for stent loading (contraction) and de-loading (expansion) under symmetric actuation of RoSE (Fig. 7D). The stent radial stiffness ($k$) can be defined as how much diameter of the stent is reduced by the application of the force exerted by the RoSE conduit. The stiffness signifies the effectiveness of the stent in resisting diameter change during RoSE contraction and expansion. By differentiating (2) with respect to $d_s$ and taking its absolute, $k$ can be given as:

$$k = \left|\frac{\mathrm{d}F_s}{\mathrm{d}d_s}\right| = \left|2\pi \frac{F_f}{W_f}\right| \tag{4}$$

For measuring the elongation strain of the stent during the pressurization cycle, a time of flight (TOF) sensor was mounted vertically at the distal side of the RoSE. By using a nylon thread through a pulley, a vertically displacing paperboard marker was connected to the distal end of the stent. The TOF sensor was used to record the displacement of the marker (stent elongation) in the vertical direction (Fig. A1 A of Supplementary Materials). Finally, the elongation strain was calculated by the change in the marker displacement (stent length) per unit stent's initial length.

The RF stent testing protocol included symmetric actuation of the RoSE layers from $L_1$ to $L_{12}$, with the same pressure simultaneously (Fig. A1 B of Supplementary Materials, Movie S1). Within the RoSE conduit, stent A and B were expanded to a diameter of 19.7 mm and 18.7 mm and then compressed to a diameter of 11.7 mm and 7.2 mm respectively.

For stent migration testing, instead of symmetric actuation, RoSE was actuated with peristaltic actuation which includes spatio-temporal sinusoidal peristalsis waves of different characteristics (Table 3, Fig. A1 C of Supplementary Materials). The migration studies elucidated the contributions of stent RF, bolus concentrations, and peristaltic wave trajectories of RoSE towards the stent migration in terms of both displacement and direction. To ensure repeatability, all the RF and migration experiments were conducted five times.

**Table 3. Manipulated parameters to cover a range of swallowing scenarios.**

| Manipulated parameters | Variations |
| --- | --- |

| Stent radial stiffness | 1.55 ± 0.24, 3.13 ± 0.53 $Nmm^{-1}$ |
|---|---|
| Bolus concentrations | 72, 108, 144 $gL^{-1}$ |
| Wave velocities | 20, 30, 40 $mms^{-1}$ [6] |
| Wavefront lengths | 40, 50, 60 mm |

*RoSE manometry protocol*

In RoSE manometry protocol, RoSE was implanted with a stent, and a manometric motility catheter (P3315205CC152, Sandhill Scientific, USA) positioned akin to the clinical in vivo observations (Fig. 8, A and B).[6] The pressure signatures, intrabolus and intraluminal, associated with each swallow was captured using the manometric motility catheter and data acquisition system (S98-200C, Sandhill Scientific, USA). The catheter is a long flexible tube with a 4 mm diameter, which consists of 5 pressure sensors evenly spaced at 50 mm. The swallow investigation was carried out by aligning one of the sensors to the layer $L_4$ of the RoSE (Fig. 8A). Due to the limited visibility inside the conduit, the configuration was achieved according to the dimension of the RoSE and the catheter. RoSE was kept in a supine position to evade the effect of gravity on the transport of the food boluses. Feeding pipes were connected at both ends of the RoSE conduit, and by using a funnel, bolus was fed to the RoSE through the left pipe.

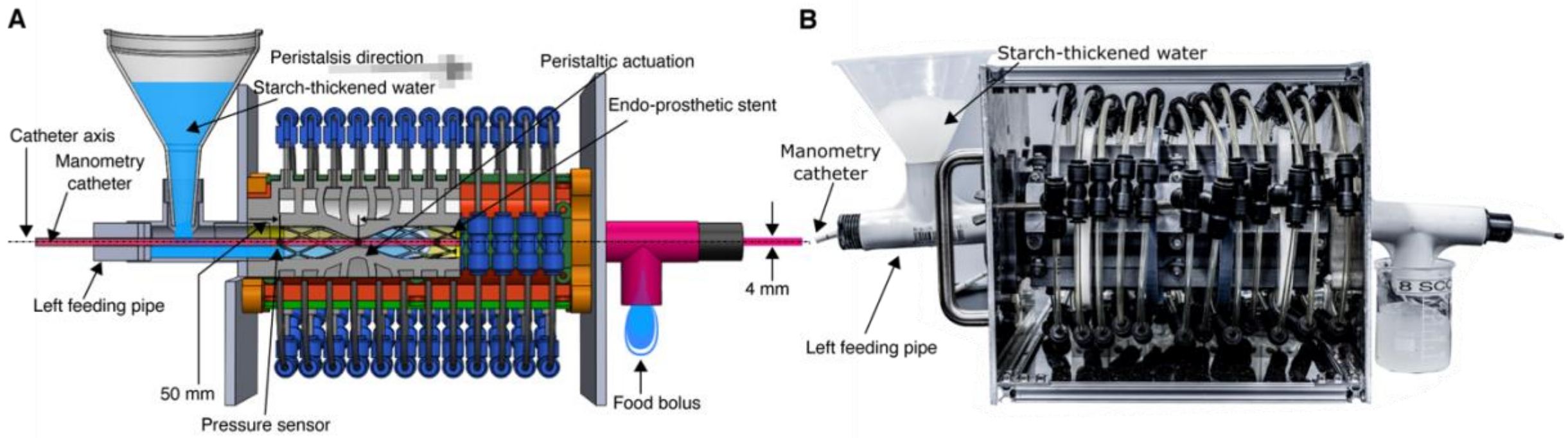

**Fig. 8. RoSE manometry test setup to measure bolus swallow pressure signatures.** (A, and B) Schematic, and image of the manometry test setup to statistically analyze the mean maximum intrabolus pressure (IBPS), and the IBPS gradient as a function of bolus concentration, peristalsis wave velocity, and stent placement (before and after).

The peristaltic actuation described in *RoSE firmware protocol* generates peristaltic waves of various wavefront lengths and velocities (Table 3). The wave generates a contractile force that squeezes and pushes the bolus from left to right of the RoSE conduit, with or without the presence of a stent (Fig. 8). The manometry protocol has been implemented to inspect the macro bolus behavior under peristalsis (Movie S2). During the bolus transport, the pressure signature recorded by the catheter pressure sensor in the presence of the bolus, near the bolus tail, is known as IBPS whereas, the intraluminal pressure signature (ILPS) is the overall contact pressure (IBPS and RoSE-catheter contact pressure) recorded along the catheter axis. The peak ILPS describes the normal contact pressure exerted by the RoSE lumen on the manometry catheter. Due to the non-Newtonian nature of the bolus fluid behavior and different peristaltic deformation characteristics owing to various wave velocities and wavefront lengths, the IBPS gradient in the region of bolus tail is an essential indicator of RoSE swallow efficacy.

The RoSE manometry protocol is designed to study the impacts of stent stiffnesses, bolus viscosities, peristaltic wave velocities, and wavefront lengths on the IBPSs (Table 3) and its gradient. The manipulated parameters (Table 3) are designed to cover a range of different swallowing scenarios, where each combination was tested in a 2 x 3 x 3 x 3 experimental protocol. Moreover, to ensure repeatability, each set of experiments was conducted five times.

## Validation Test Results

### *Stent RF analysis with RoSE*

To mitigate migration, the stent designing criteria is to maintain a high RRF and a low COF as possible.[11, 12, 19] To explore the stent RF response as a function of its diameter, FSP was implanted in RoSE, followed by subsequent deployment of two silicone covered SEMS: stent A and B (Fig. 6), with mean radial stiffness of 1.55 ± 0.24, 3.13 ± 0.53 $Nmm^{-1}$ respectively (Fig. 7, Movie S1).

With recurring simultaneous inflation and deflation of the RoSE air chambers in each layer ($L_1$, $L_2$, ..., $L_{12}$), the observations on the stent RF plotted an interesting hysteresis curve depicting the measure of RRF, and the COF (Fig. 9 A, and B). This pattern of hysteresis, permitted the COF exerted on the RoSE (esophageal) wall to remain low during the large deflections as well as the radial oversizing of the stent which is caused by factors external to the stent geometry and its functions. On the other hand, it could be witnessed that the stent resistance to compression, the RRF rose with the deflection (austenite phase) until the RF plateaued (martensite phase). Similar kind of hysteresis have been indicated in other studies.[11, 12, 40].

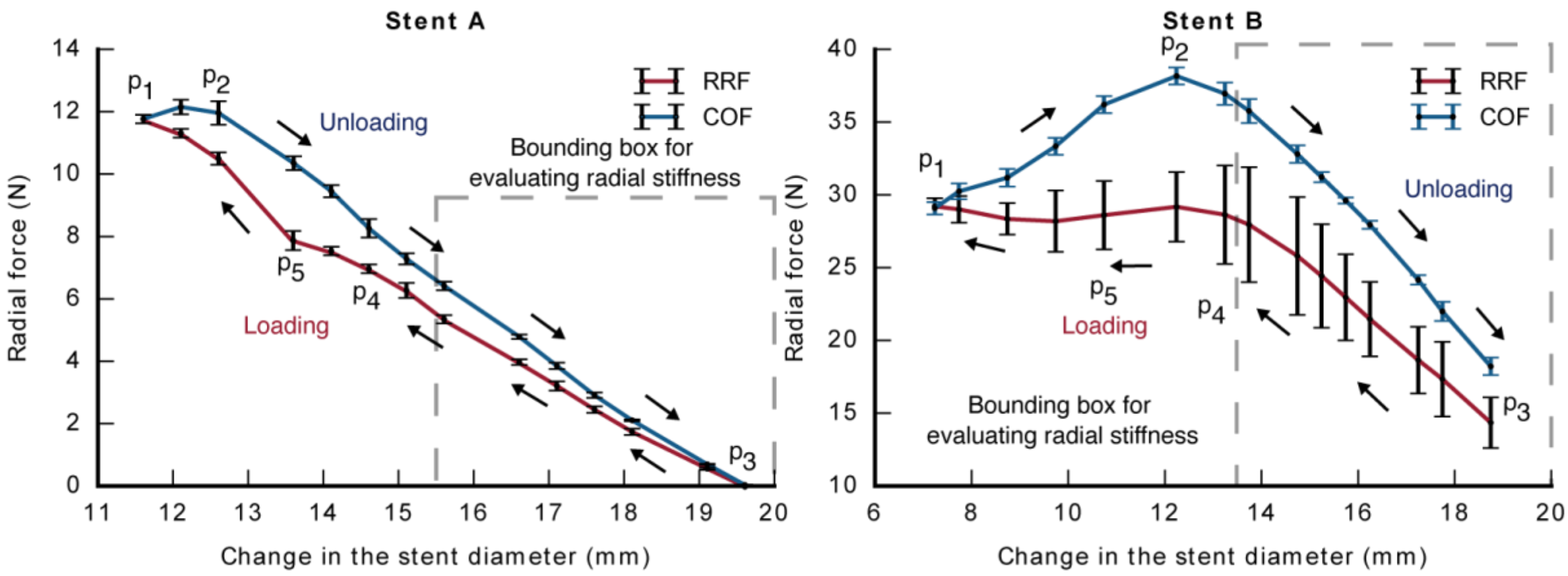

**Fig. 9. RF testing results.** (A, and B) Plots showing the hysteresis in the RF as a function of the stent diameter for stent A and B, respectively. For stent A and B, the initial value of the COF (chronic outward force) is around 12.06 ± 0.14 and 29.01 ± 0.42 N ($p_1$) which gradually increased for deflations up to diameters of 12.4 and 12.2 mm ($p_2$), respectively. On further deflation of the RoSE up to 19.7 and 18.7 mm, respectively ($p_3$) for stent A and B, the COF values exhibited around a linear response with a negative slope for both the stents. However, under the fully distended state of the RoSE ($p_3$), stent A has shown a relatively low expansive RF of 0.33 ± 0.01 N (<15 N negative value), and stent B has recorded a value of 18.22 ± 0.60 N for the same. The RRF (radial resistance force), response for stent A to the periodic expansion and contraction is, however, outside a linear definition and instead exhibited a two-way accession. Upon contracting, until 15.7 mm diameter ($p_4$), the RRF linearly increased to 5.65 ± 0.13 N with a gradient of 1.37 ± 0.05 $Nmm^{-1}$, and after that, transformed into the martensite phase where the force gradient reduced. Finally, at the end of compression ($p_5$), the RRF accounted for 12.06 ± 0.14 N with a gradient of 0.9 ± 0.01 $Nmm^{-1}$. For stent B, the RRF increased rapidly with the contraction in the RoSE, until the force curve plateaued at 27.95 ± 3.95 N.

In Fig. 8, the COF recorded for stent A, and B gradually increased from $p_1$ to $p_2$. After $p_2$, both the plots have displayed a fast linear decrease in the COF up to $p_3$. At $p_3$, stent A has shown a relatively low expansive RF (COF) of 0.33 ± 0.01 N (Fig. 9A), and stent B has recorded a value of 18.22 ± 0.60 N for the same (Fig. 8B). For RRF, both the plots have shown a rapid increase in the region between $p_3$, and $p_4$. In stent A, upon contracting until $p_4$, the RRF linearly increased and after that, transformed into the martensite phase ($p_4$ to $p_5$), where the gradient reduced (Fig. 9A). In stent B, from $p_4$ to $p_5$, the RRF plateaued drastically, taking the force gradient close to zero (Fig. 9 B). The stent A and B radial stiffness were calculated by taking the mean of the slope at each point within the bounding box (linear region of COF and RRF) in Fig. 9. The mean radial stiffness for stent A and B were found to be 1.55 ± 0.24 $Nmm^{-1}$ and 3.13 ± 0.53 $Nmm^{-1}$. Given the Poisson effect, stent A and B exhibited 56.60 ± 2 and 17.70 ± 2 % of elongation in its length, respectively, during contraction with an applied chamber pressure of 50 ± 0.35 KPa (Fig. A2 of Supplementary Materials). The comparison of the stiffer stent B with stent A revealed that its overall RF is twice as higher as compared to stent A. A relatively higher variation in the elongation of stent A maintains its RF at a low profile.

*RoSE manometry during regular stent operation*

The RoSE manometry reading represents a measure of the bolus pressure variation in response to the active contractile forces originating in its wall (Fig. 8). A portion of the contractile force goes into accelerating the bolus, and a part goes into accelerating the wall. In contrast with the fixed time spatial pressure variation through a bolus, manometric recordings within the lumen provide the pressure as a function of time at fixed spatial points (Fig. 10A).[6] When the bolus is in motion, the output of the manometric pressure transducer in the bolus presence region is the hydrodynamic pressure exerted by the bolus, which is the IBPS (Fig. 8A, and Fig. 10A) and is significant for food innovation studies. The IBPS exhibits rapid variation in the region of high frictional forces (e.g., bolus tail region), which leads to the occurrence of friction-induced IBPS gradient.[6] The IBPS gradient in the region of the bolus tail is a critical indicator of swallow efficacy. The tip of the bolus tail creates a demarcation between the region of IBPS and the ILPS. As long as the peak ILPS is satisfactory to maintain a wave seal behind the bolus tail; its magnitude is unimportant in terms of successful bolus transport.[37] A peak ILPS of at least 16 KPa (120 mmHg) was achieved in all the manometry experiments, which is higher than the clinical measurements, and hence it is acceptable. [6]

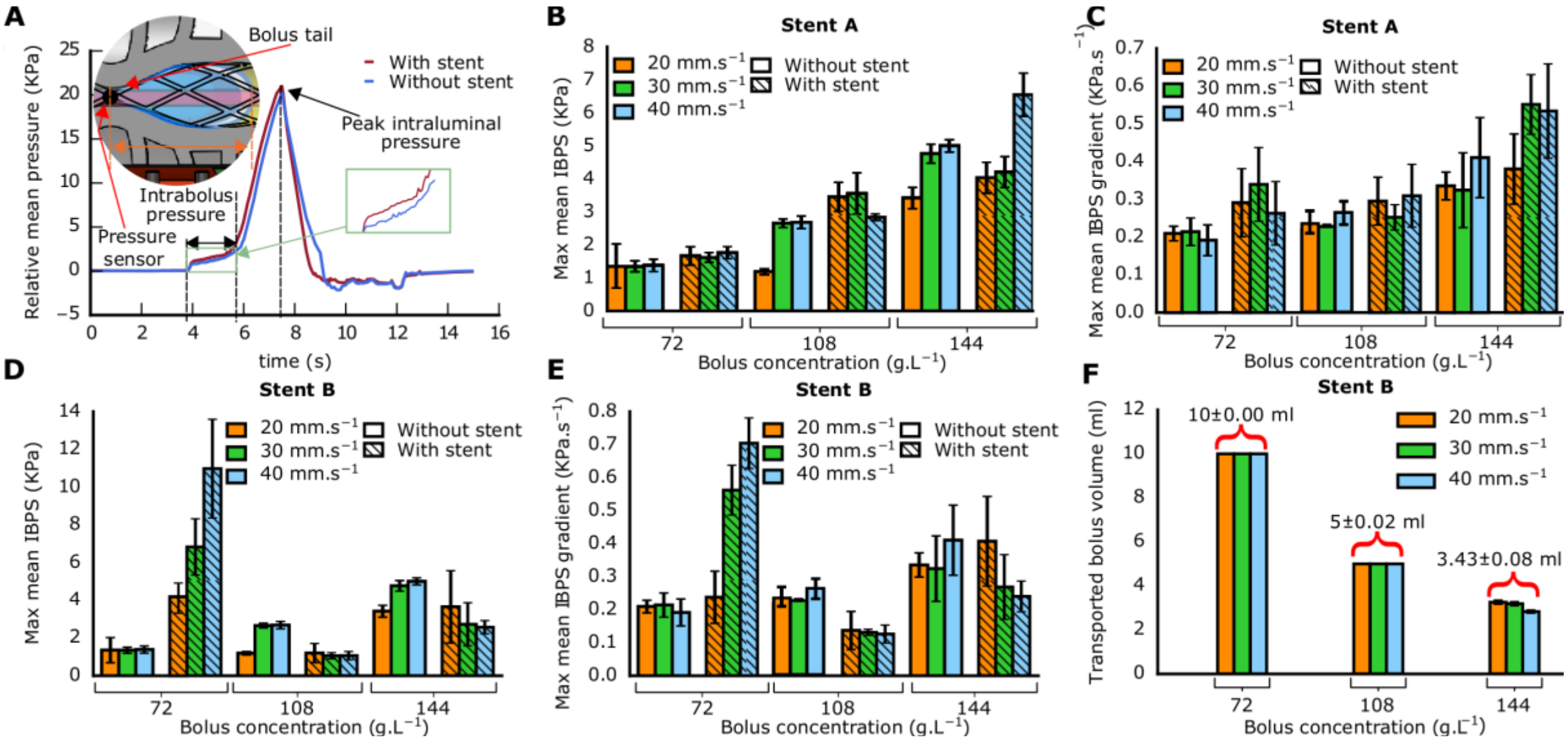

**Fig. 10. Manometry testing results.** (A) Comparison of the plots, showing the relative mean pressure with time in presence and absence of stent A, recorded by the manometry catheter pressure sensor #2 coinciding with the layer $L_4$ of the RoSE. The plot also depicts the intrabolus pressure (IBPS), and peak intraluminal pressure. In this study, IBPS refers to the maximum mean IBPS. (B, and C) Plots showing the IBPS and its gradient measured at the bolus tail with and without stent A. The data was generated by actuating the robot with peristalsis velocities of 20, 30, and 40 mms$^{-1}$. Increase in the IBPS and its gradient were mostly affected by the bolus concentration and stent A implantation and less so by the wave velocity. With stent A implanted, from bolus I to III, the IBPS increased for all peristalsis velocities. For peristalsis velocity of 40 mms$^{-1}$ and bolus III, the IBPS and the gradient rose from 5 ± 0.18 to 6.54 ± 0.65 KPa, and 0.41 ± 0.10 to 0.53 ± 0.12 KPa.s$^{-1}$ respectively, when stent A was deployed. (D, and E) Plots showing the IBPS and its gradient measured at the bolus tail with and without stent B. The reduction in the IBPS and its gradient was mostly affected by buckling of stent B, and less so by the wave velocity. After the deployment of stent B, the initial fall in the IBPS and its gradient ([0.24 ± 0.08, 0.56 ± 0.07, 0.70 ± 0.08] to [0.14 ± 0.06, 0.13 ± 0.01, 0.13 ± 0.03] KPa.s$^{-1}$) from 72 to 108 gL$^{-1}$ is due to the reduction in transported bolus volume. From 108 to 144 gL$^{-1}$, the rise of IBPS and IBPS gradient ([0.14 ± 0.06, 0.13 ± 0.01, 0.13 ± 0.03] to [0.40 ± 0.14, 0.27 ± 0.10, 0.24 ± 0.05]) was due to the increase in bolus viscosity (F) Plot illustrating the variation of the transported bolus volume in the presence of stent B. Due to the buckling of stent B after implantation, a reduction in the transported bolus volume was recorded with the increase in the bolus concentration for all wave speeds. For bolus I, II, and III, mean transported bolus volume was determined by averaging the volume of bolus transported for each peristalsis velocity.

In the bolus tail region, the IBPS (IBPS refers to the maximum mean IBPS) rose sharply even for a slight reduction in the conduit diameter. During the peristaltic transport of the bolus, the IBPS is highly profound to the intraluminal geometry in the greatly occluded region (especially near the bolus tail tip). With the presence of stent A, in such a region, the IBPS pressure distribution becomes relatively higher because the stent's braided structure slightly reduced the intraluminal geometry (Fig. 10, A, and B). This finding corroborates the ideas of Brasseur et al.[6], who suggested that during peristaltic bolus transport, the IBPS is extremely sensitive to the geometry of the intraluminal, in the profoundly occluded regions. For low starch concentration bolus I, the IBPS have shown a slight increase in the presence of stent A but, for the higher starch concentration bolus III, the IBPS has increased significantly, when stent A was deployed for all velocities (Fig. 10B).

The findings of the manometry experiments in the presence and absence of stent A indicated that higher concentration starch boluses have exhibited risen pressure gradient in the wave tail (Fig. 8C). From bolus I (72 gL $^{-1}$) to III (144 gL$^{-1}$), a significant rise in the pressure gradient bars was exhibited. A substantial increase in the pressure gradient was observed from bolus II to III. The findings are consistent with the results of the rheological investigation conducted by Dirven et al.[37] Mainly, the highest wave speeds generated the highest-pressure gradient bars (Fig. 10C). Following the addition of stent A, resulted in high frictional forces in the contractile zone of the RoSE conduit and hence, a significant increase in the gradient was recorded for all types of bolus formulations, and velocities (Fig. 10C). The result match with the findings of Brasseur et al.[6] The author proposed that the frictional force induced IBPS gradient is inversely proportional to the fourth power of the intraluminal cross-sectional diameter of the esophagus. With the deployment of the stent A in the RoSE conduit, the conduit diameter has slightly reduced, and thus, the IBPS gradient rose more sharply as compared to the no stent condition. The findings of the experiments have not exhibited significant change in the IBPS and the IBPS gradient with respect to wavefront length, since the variation in wavefront does not predominantly vary the wave tail wall gradient.

*RoSE manometry during stent dysfunctionality*

A stiffer stent like stent B significantly decreased the radial compliance of the lumen by introducing buckling in it. Buckling would lead to inadequate apposition of the stent in the conduit wall and reduce its ability to transport higher starch concentration boluses by acting as an obstacle in the path of bolus transport (Fig. 11). The buckling can cause stent dysfunction over time, which could cause recurrent dysphagia and, consequently, food bolus obstruction,[41, 42] and swallow inefficacy. Once the stent buckles, its stiffness is generally dramatically reduced, which may eventually cause stent coating defects or a fracture or both.[22] Any structural dysfunctionality in the stent can snag the food bolus.[42] The patterned coating on stent B (Fig. 6B), maximizing its radial stiffness, made it susceptible to buckling around its flares (since the flare diameter is higher than the body diameter) during its deployment.

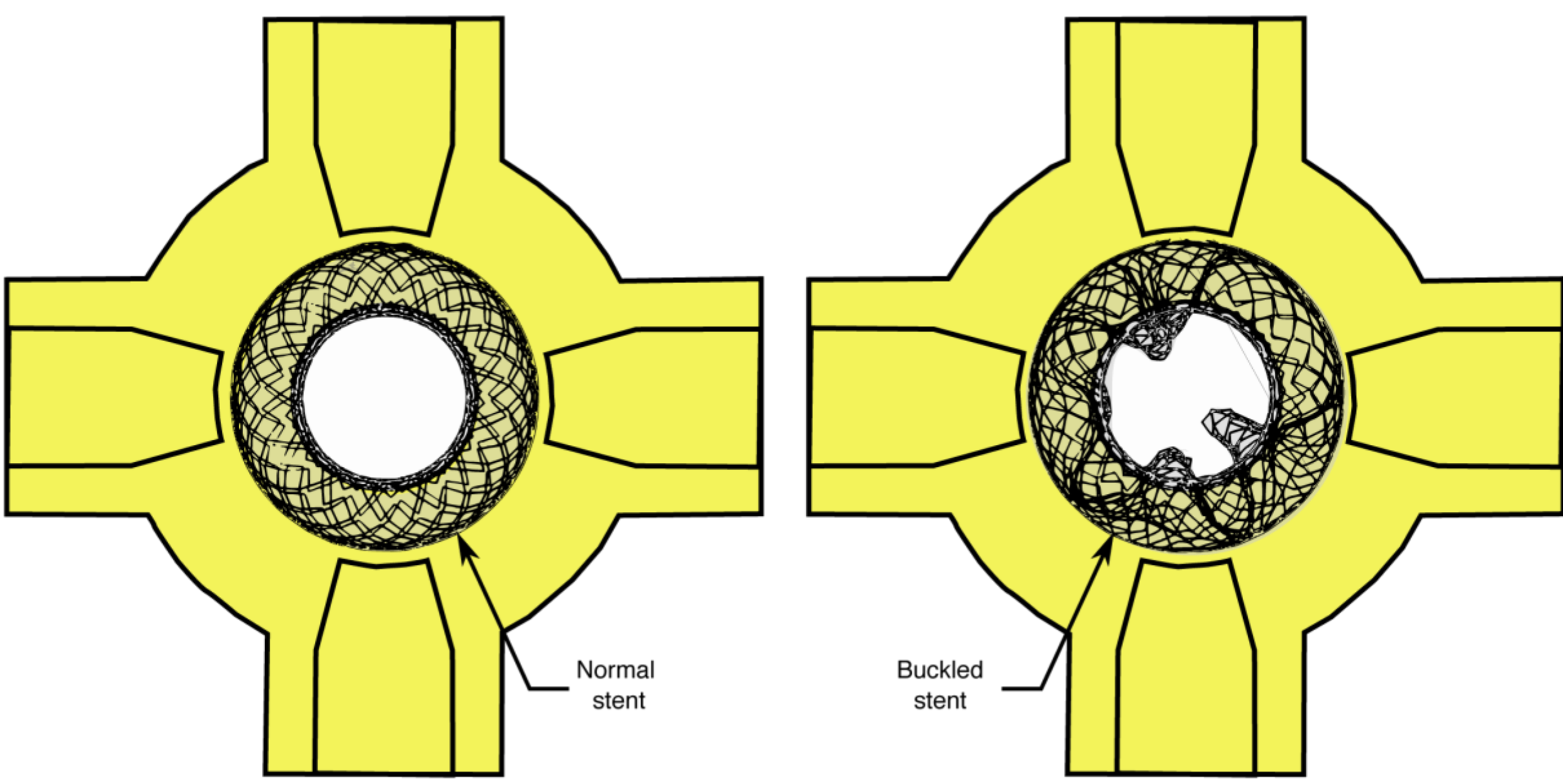

**Fig. 11. Illustration of implanted stents in RoSE, normal (left) and buckled (right).** Buckling refers to the stent unstable deformation outside the circumferential plane due to the circumferential compression. A stiffer stent (like stent B) can reduce the radial compliance of RoSE conduit by buckling.

The actuation trajectory propagated the bolus forward, but the buckled stent B posed an obstacle during transportation, resulted in bolus fluid reflux, and reduction in the transported bolus volume. The phenomenon became prominent for higher viscous bolus II and III with higher velocities (30 and 40 $mms^{-1}$), which resulted in the decrease of IBPS and its gradient (Fig. 10, D, and E). After a bolus concentration of 72 $gL^{-1}$, a significant difference between the IBPS (and IBPS gradient) has been recorded in the presence and absence of stent B. From bolus I to II, IBPS, and IBPS gradient drastically dropped (Fig. 10, D, and E). The buckling of stent B has restricted the mean bolus II, and III volume to 5 ± 0.02, and 3.43 ± 0.08 ml per cycle, respectively (Fig. 10F). In the case of bolus II, the reduction in the IBPS is overwhelmed by the bolus volume, and not by the bolus viscosity. The dependency of the IBPS on the viscosity is only applicable for a minimum bolus volume of 10 ml. Similar results have been reported by Ren et al.[43] Interestingly, in contrast to the mean bolus II volume (5 ± 0.02 ml), although the mean bolus III volume (3.43 ± 0.08 ml) decreased, an increase in the IBPS and IBPS gradient was recorded due to the rise in the bolus III viscosity (Fig. 10, D, and E).

In general, the IBPS, IBPS gradient, and the transported mean bolus volume obtained after implantation of the stiffer stent B suggested that the stent dysfunctionality reduced the swallow efficacy from bolus I to II. In addition, the analysis also suggested that the swallow efficacy can be improved by further increasing the bolus viscosity to bolus III. The RoSE manometry results inspected the macro bolus behavior under peristalsis (Movie S2), concluded the interactions between viscosity and peristaltic parameters (Table 3), and predicted the swallow efficacy before and after stent implantation.

*Stent migration testing*

Understanding the physiology of swallowing has led us to hypothesize that axial displacement of the esophageal wall (RoSE conduit wall) is a significant factor in stent migration. The migration of a stent is correlated with the RF developed due to the contact pressure of the stent (COF divided by the contact area of the stent). The migration tests were done for dry (no bolus) and bolus swallow (bolus I, II, and III) conditions (Table 2). The measured migration displacement is relative and not absolute to the initial stent deployment position. The relative positive (negative) displacement values indicated the migration similar (opposite) to the direction of the peristalsis (Fig. 12).

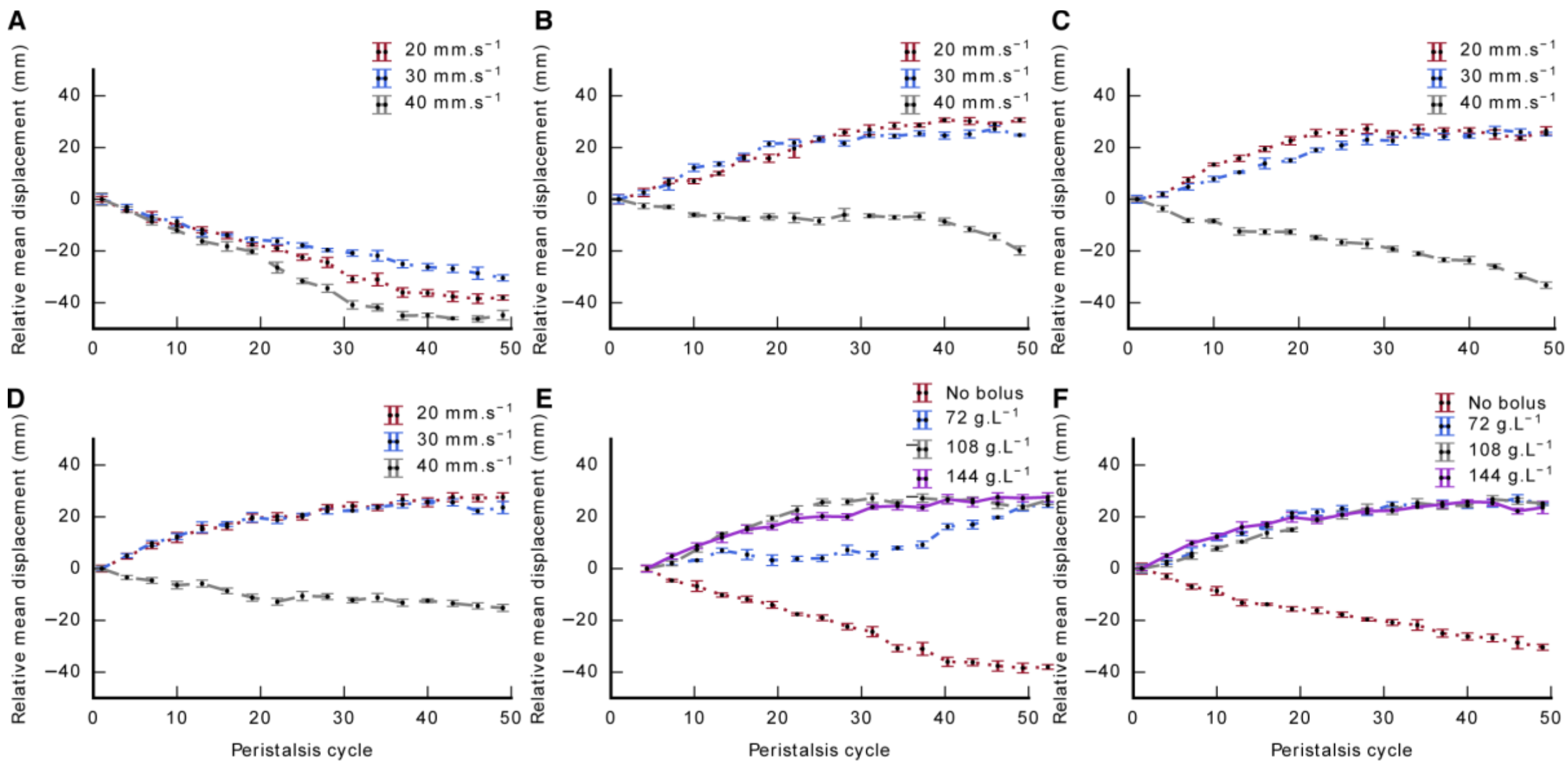

**Fig. 12. Relative stent mean migration displacement results as a function of the number of peristalsis cycle.** Plot(s) showing the relative mean migration displacement of stent A, (A) in dry swallow conditions (without bolus). The stent migrated contrary to the direction of the peristalsis, thereby causing a negative relative displacement. (B to D) with bolus I, II, and III (72, 104, and 144 $gL^{-1}$), respectively. Since the net contractile force in the direction of the peristalsis was higher than the trajectory induced frictional force (TIFF), the stent moved in the direction of the peristalsis for 20, and 30 $mms^{-1}$. For 40 $mms^{-1}$, the stent has shown a negative displacement due to the lower contractile force as compared to the TIFF. (E, F) With the increase in bolus concentration, the results have shown mostly a slight increase in the migration displacement for all velocities.

Stent A and B exhibited intriguing variations when tested for migration. The maximum relative displacement of 47 ± 1.7 mm for 50 peristalsis cycles was recorded for stent A (Fig. 12A), whereas, for stent B, the inspected measurements were negligible. Migration was commonly recorded in stent A, which was caused by the insufficient RF exerted by the stent. For dry swallow experiments, the stent moved contrary to the direction of the peristalsis for all peristalsis velocities (Fig. 12A) while, the results obtained from the bolus swallow have shown the stent migration in the direction of the peristalsis (Fig. 12, B to D)

The cause of the negative displacement can be well explained from the marker trajectory data collected from the quarter version of the RoSE experiments (Fig. 13, B to E, Movie S3). The quarter RoSE models the one-fourth radial cut of the RoSE along its axis such that the model retains one chamber per layer along the RoSE length (Fig. 13, A, and B).[39] The quarter RoSE is developed to address the issue of limited visibility inside the conduit of the RoSE. As the peristalsis trajectory admitted, the z-axis marker displacement increased with the initial decrease in x-axis values (axial displacement), creating a displacement trajectory (Fig. 13F). This initial reduction has provided evidence to the fact that any point on the RoSE surface moved opposite to the direction of the peristalsis, thereby inducing a frictional force on the stent in the peristalsis direction (Fig. 13, C to E, Movie S3). This frictional force is defined as the trajectory induced frictional force (TIFF).

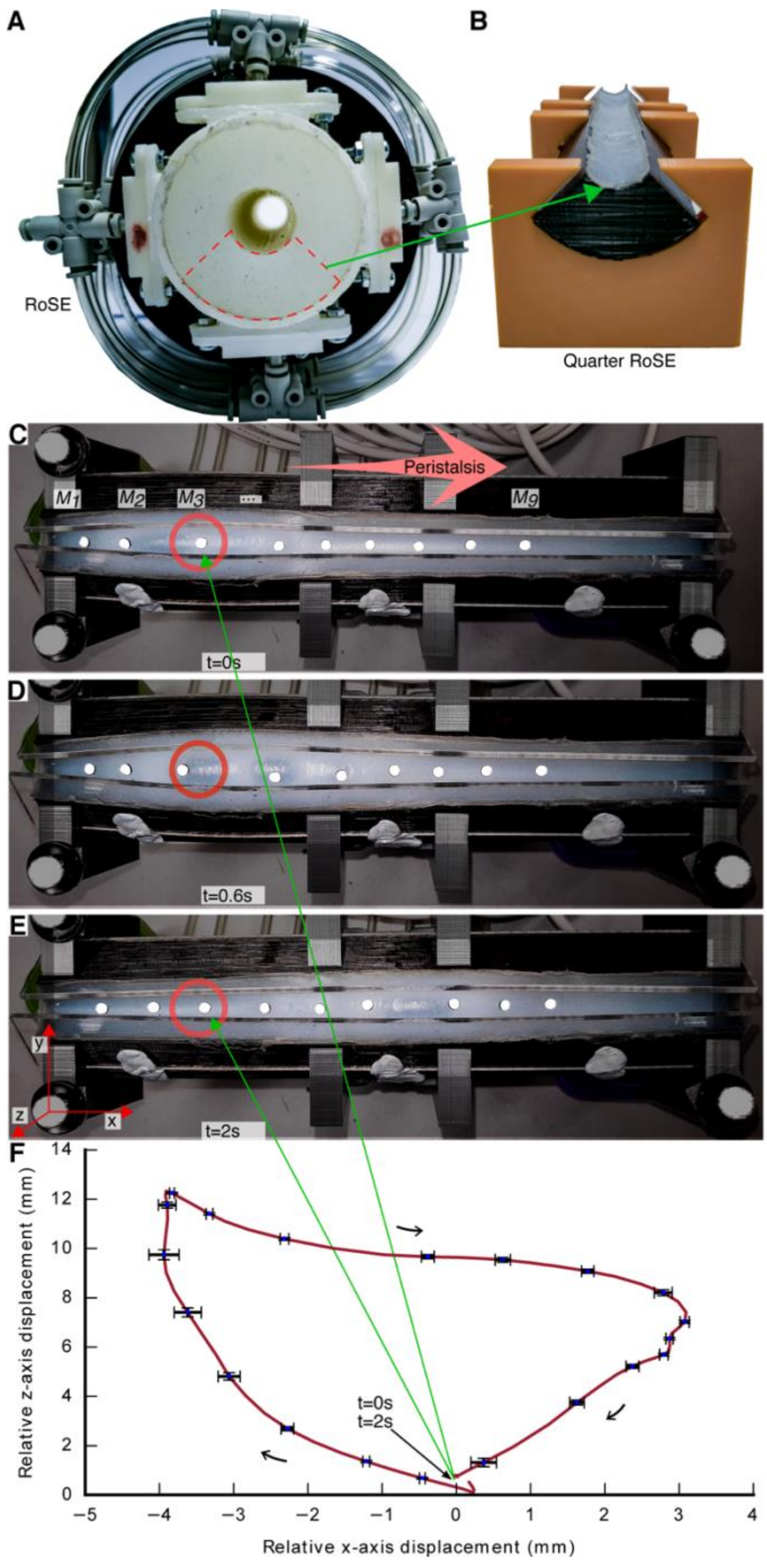

**Fig. 13. Demonstration of peristaltic actuation in a quarter RoSE.** (A to B) A quarter version of the RoSE is developed to visualize the peristaltic motion inside the RoSE conduit. (C to E) A set of 9 markers ($M_1$, $M_2$, … $M_9$) were attached on top of each layer of the quarter RoSE, and time-series x-, z- movement data of the markers were recorded.[50] (E) Plot showing the x-, z- displacement trajectory of the marker $M_3$ (encircled in red). With the commencement of the peristalsis in the quarter RoSE towards the right, the relative z-axis displacement of $M_3$

increased while its relative x-axis displacement reduced. The characteristics of the plotted trajectory indicated that any point on the RoSE surface moved opposite to the direction of the peristalsis.

With the introduction of the synthetic boluses in the RoSE lumen, the stent commenced movement in the peristalsis direction for velocities of 20 and 30 $mms^{-1}$ (Fig. 12, B to D). Since the stent RF was not very high, a tiny fraction of the bolus fluid was able to leak in between the RoSE layer and the outer stent surface. As a result, the TIFF reduced, and the prolonged peristaltic contractile force in the direction of the peristalsis was able to overcome the TIFF. For 40 $mms^{-1}$, a similar phenomenon occurred, but the contractile force remained lower than the TIFF (Fig. 9). For all peristalsis velocities, the results mostly exhibited a slight elevation in the relative migration with an increase in bolus starch concentration (Fig. 12, E, and F).

The migration measurements for stent B were negligible. In the resting state of RoSE, stent B has reported a high expansive RF (COF) of 18.22 ± 0.60 N (Fig. 9B) as compared to stent A (0.33 ± 0.006 N, Fig. 9A). High COF of stent B ensures proper fixation of the stent on the RoSE conduit wall, under peristalsis, preventing stent migration. The results have also been confirmed by Isayama et al.[12] Taken together, the findings further confirmed that stent migration is associated with lesser RF. Further, a relatively high RF of the stent is required to guarantee its *in situ* position on the RoSE conduit to prevent stent migration and maintain adequate luminal patency. This could ensure its performance when deployed in the event of an esophageal stricture.

**Discussion**

Palliative therapy for dysphagia caused due to benign and malignant esophageal strictures can be approached with the implantation of endoprosthetic stents.[2, 8] The stenting can provide the patients with instant relief, and it can restore their oral nutrition requirements in one session (Fig. 1). However, adverse events like stent migration and various stent dysfunctionalities present vital obstacles to successful stent deployment in the esophagus.[13] Due to the less known relationship between the mechanical properties of the esophageal stents and their clinical outcome, the stenting guidelines are still poorly defined, and it relies solely on the experience of the endoscopists. The existence of little evidence from randomized controlled trials in patients restricts the comparison of different stent designs to a limited range.[16, 44]

According to the US FDA guidelines for SEMS, compression force, expansion force, corrosion, tensile strength, deployment, and dimensional testing need to be performed on the stents to establish their substantial equivalence.[18] Many physical attributes of RoSE, such as conduit length, diameter, peristalsis velocity, and conduit wall activation, are inspired by the physiological features of the human esophagus (Table 1). Hence, stents of various coatings (SEPSs, SEMSs, and biodegradable stents), cover patterns, and flare shapes can be readily deployed in RoSE. Besides, one of the significant qualitative characteristics of RoSE is its compliance, which makes it a very suitable platform for deploying stents of various characteristics. In contrast, RoSE has a stack of regular and repeating layers of pneumatic hollow chambers, arranged axis-symmetrically (Fig. 2); hence, by controlling the air pressure in each layer, RoSE conduit diameter can be varied over a range of 0 to 20 mm. The feature of RoSE, as mentioned earlier, makes it a perfect choice for conducting compression and expansion stent testing. Additionally, RoSE mimics the human swallowing action; thus, RoSE is capable of generating peristaltic waves, which is a requirement for studying stent migration.

In this study, to validate the application of biomimetic RoSE in proving substantial equivalence of esophageal stents of different characteristics, compression, and force testing of stent A and B with different radial stiffness have been performed (Fig. 7). This study has proposed a novel approach to study the mechanical behavior of SEMSs implanted in a biomimetic RoSE, under several artificial peristaltic conditions (Table 3). The ability of the RoSE to generate various peristaltic wave conditions is crucial for the conducted stent migration and bolus swallow experiments. The independent peristaltic variables allowed the study to fit larger pathological models with high inter-swallow reliability. The validation test results have proved RoSE to be a novel *in-vitro* stent testing platform.

The study has also validated the applicability of RoSE on assessing the influence of the stent RF on stent migration, and the impact of stent stiffness on swallow efficacy. Stent A and B with lower and higher stiffnesses were deployed in the RoSE subsequently, to analyze the effect of RF on the stent migration (Fig. 12). The relatively lower radial stiffness of stent A has maintained its RF at a low profile while higher RF in stent B has been associated with higher stiffness. The implication of the results suggested that higher RF of the stent is required to minimize the stent migration under the prolonged peristaltic contractile forces in the RoSE and the esophagus. Higher RF of the stent also ensures its performance when deployed in the event of an esophageal stricture.

In an unimpaired human swallowing process, the same peristaltic wave propagates throughout the esophagus, resulting in complete occlusion along its length. Hence, the stent undergoes similar deformation irrespective of its location of placement. In this study, we have assumed that active deformation of the esophagus primarily drives stent migration during swallowing. Further, the peristaltic wave in human swallowing speed is relatively faster in the middle of the esophagus as compared to the upper or lower part. By varying the peristaltic wave speed in RoSE, the range of speed for upper, lower, and middle esophagus have been achieved. The effect of speed variation on stent migration is shown in Fig. 12. Based on the theory of virtual work by Jedwab and Clerc [20], it can be concluded that the radial pressure applied by the stent on the RoSE wall is independent of the wall elastic properties. Thus by considering uniform crimping throughout the RoSE conduit and negligible difference in stent penetration depth among the different muscle fibres, stent RF remains the same irrespective of different muscle fiber region in the esophagus. Still, further studies needs to be done to evaluate the interaction of stents with different esophagus muscle fibre regions.

To mimic the swallowing physiology and for evaluating swallowing efficacies, synthetic bolus formulations of different starch concentrations and viscosities (Table 2) were tested on the RoSE, with or without the stent implanted. The RoSE manometric study has allowed us to evaluate the IBPS and its gradient under different bolus and stenting conditions. For stent A, it was found that higher bolus concentration significantly raised the IBPS and its gradient. With the stent implanted, they further enhanced (Fig. 10, B and C). A stiffer stent like the stent B significantly reduced the radial compliance of the RoSE conduit, and thus buckled inside it. The buckling acts as an obstacle in the path of successful bolus transport, which became more prominent for higher starch concentration boluses, and thereby significantly decreased the IBPS, and its gradient (Fig. 10, D to F) leading to swallowing inefficacy. Thus in the area of stent design and testing, RoSE can provide an innovative platform for testing endoprosthetic stents of various designs to address numerous clinical challenges associated with successful stent implantation. However, the need for high-quality clinical data from randomized controlled trials (RCTs) and the fact that more factors are probably involved in clinical stent behavior makes it challenging to correlate the manifested results to clinical outcomes effectively. In addition to the RF, the axial force, stent cover and wire material, wire diameter, angle of wire crossover, may also play a significant role in the clinical outcome. The evidence from RCTs that can be used to test the clinical performance of the stent based on the RF and other parameters, as mentioned earlier, is little.

*Limitations*

While the study is comprehensive, there are a few limitations that are worthy of being mentioned. All the tests mentioned in the US FDA guidelines for SEMS can be performed on RoSE but, in this study, the capability of the RoSE is limited to expansion and compression force testing of the stents.[18] However, some corrosion in the stents were observed due to their interaction with the artificial saliva and the food boluses (artificial starch-thickened water). Corrosion and tensile strength testing of the stents on the RoSE were not performed. Although FSP offered a proper measurement of the RF, it can only be operated up to a specific axial bending range. The reason for using FSP was due to the unavailability of any embedded pressure sensor in the RoSE. During the stent RF testing, although the stents were allowed to move freely, the contraction of RoSE outside the footprint of the stents limited their elongation up to some extent. In addition, the intrinsic hysteresis of the RoSE might have some contribution to the observed hysteresis in Fig. 7.

Even though RoSE mimics the esophagus in various physiological ways but its conduit still lacks artificial strictures. Mainly, the endoscopic stents are implanted in the esophagus with benign and malignant strictures. Also, due to the absence of a closed-looped controller, the peristaltic contractions in the RoSE conduit rely on the predefined trajectories, and the trajectories are not adaptive to the rheology of the food bolus. Besides, the entire RoSE conduit is made up of the same silicone rubber material (Ecoflex 0030, Smooth-on, USA), while the muscles in the esophagus are distributed into striated and smooth circular muscle fibers with different actuation patterns.[35]

## Conclusion

Benign and malignant esophageal strictures would cause swallowing impairment (dysphagia) in patients who have esophageal cancer. Endoprosthetic stent placement can provide immediate and cost-effective therapy for patients suffering from dysphagia caused by esophageal cancer. After the procedure, the implanted stents in the patients are often associated with an inadequacy known as stent migration, weakening their swallowing efficacy. The stent RF on the esophageal wall is a crucial stent design parameter to minimize stent migration after the implantation. However, due to limited randomized controlled trials in patients, the stent design and stenting guidelines are still unconstructive. Hence, for in vitro testing of endoprosthetic stents, a bio-mimicking RoSE was developed.

The primary focus of this study was to determine whether RoSE is a suitable platform for performing different stent-related measurements and testings. Experimental validation tests of stent A and B were carried out to prove the application of RoSE in the field of stent testing. The first two objectives of the study were to measure stent RF and migration under various peristaltic conditions. To conduct the experiments, stent A and B were implanted in the RoSE to measure their respective RFs and the impact of RF on their migration. Both the objectives were met successfully. To study the effect of the stent and its dysfunctionality on the IBPS in the RoSE, and further its impact on the swallowing efficacy, endoscopic manometry tests were performed on the RoSE with different peristalsis trajectories. By introducing artificial saliva, synthetic boluses of starch thickened water, and implantation of stent A and B, experiments for the final two objectives were performed. The results of the experiments have shown the achievement of the goals.

*Future Scope*

In future work, artificial strictures in RoSE, of various shapes and sizes will be fabricated by using Ecoflex 0030, which is also used for constructing the RoSE conduit (Ecoflex 0030, Smooth-on, USA). Besides, by employing different silicone rubber materials, striated, and smooth muscle regions in RoSE can be created. In the future, RoSE will be devised to study the interaction of the fifteen stents (including stent A and B) mentioned in *Stent configuration*, with strictures of different shapes and sizes. Besides, the limitation of the elongation of the stent along its axis, due to RoSE contraction outside its contact area, can be addressed by selecting a pressure profile which can only pressurize the layers adjacent to the stent. To validate the RoSE-stent interaction results for stent RF and migration, finite element model (FEM) will be considered in future. The model will be similar to the FEM developed by Garbey et al.[15] Further studies that take stent corrosion and dimensional testing into account will be conducted, under artificial saliva (Aquae Dry Mouth Spray, Hamilton) and bolus I, II, and III (prepared from Altrix Rapid Thickener, Douglas Nutrition Ltd, New Zealand). RoSE with embedded sensing capabilities will enhance the measurement of stent RF, IBPS, and migration. A flexible and stretchable capacitive sensor, made up of carbon black is under the design process, to measure radial pressure and conduit strain. Besides, work on implementing a closed-loop controller for RoSE for controlling the deformation of the conduit is already undergoing.

RoSE in vitro stent testing can be used by the endoscopic industries to test their new stent designs in terms of migration and the occurrence of any kind of dysfunctionality during and after the stent implantation. Besides, RoSE can be used as a tool for a deeper understanding of the human impaired swallowing behavior with and without any stent implanted.

## Acknowledgements

D.B. and S.J. acknowledge doctoral scholarships from the Riddet Institute, a centre of research excellence, New Zealand. D.B. also acknowledges his doctoral scholarship from the University of Auckland, New Zealand.

## References

1. Dodds WJ. The physiology of swallowing. Dysphagia. 1989;3(4):171-8. doi: 10.1007/BF02407219.
2. Groher ME, Crary MA. Dysphagia: clinical management in adults and children. 2nd ed. St. Louis, Missouri: Elsevier; 2015.
3. Garcia JM, Chambers Iv E, Clark M, Helverson J, Matta Z. Quality of care issues for dysphagia: modifications involving oral fluids. Journal of Clinical Nursing. 2010;19(11-12):1618-24. doi: 10.1111/j.1365-2702.2009.03009.x.
4. Kuo P, Holloway RH, Nguyen NQ. Current and future techniques in the evaluation of dysphagia. Journal of Gastroenterology and Hepatology. 2012;27(5):873-81. doi: 10.1111/j.1440-1746.2012.07097.x.

5. Cichero JA, Lam P, Steele CM, Hanson B, Chen J, Dantas RO, et al. Development of international terminology and definitions for texture-modified foods and thickened fluids used in dysphagia management: The IDDSI framework. Dysphagia. 2017;32(2):293-314. doi: 10.1007%2Fs00455-016-9758-y.
6. Brasseur JG, Dodds WJ. Interpretation of intraluminal manometric measurements in terms of swallowing mechanics. Dysphagia. 1991;6(2):100-19. doi: 10.1007/BF02493487.
7. Roy N, Stemple J, Merrill RM, Thomas L. Dysphagia in the elderly: Preliminary evidence of prevalence, risk factors, and socioemotional effects. Annals of Otology, Rhinology & Laryngology. 2007;116(11):858-65. doi: 10.1177/000348940711601112.
8. Elhanafi S, Said S, Cooper C, Alkhateeb H, Othman M, McCallum R. Esophageal perforation post pneumatic dilation for achalasia managed by esophageal stenting. The American Journal Of Case Reports. 2013;14:532-5. doi: 10.12659/AJCR.889637.
9. Dua KS. Esophageal Stenting for Relief of Dysphagia. In: Belafsky Peter C, Postma, Gregory N, Easterling C, editors. Principles of Deglutition. New York, NY: Springer;2013. p. 877-888.
10. Hanawa T. Materials for metallic stents. Journal of Artificial Organs. 2009;12(2):73-9. doi: 10.1007/s10047-008-0456-x.
11. Hirdes MM, Vleggaar FP, De Beule M, Siersema PD. In vitro evaluation of the radial and axial force of self-expanding esophageal stents. Endoscopy. 2013;45(12):9971005. doi: 10.1055/s-0033-1344985.
12. Isayama H, Nakai Y, Toyokawa Y, Togawa O, Gon C, Ito Y, et al. Measurement of radial and axial forces of biliary self-expandable metallic stents. Gastrointestinal endoscopy. 2009;70(1):37-44. doi: 10.1016/j.gie.2008.09.032.
13. Sharma P, Kozarek R. Role of esophageal stents in benign and malignant diseases. The American journal of gastroenterology. 2010;105(2):258-73. doi: 10.1038/ajg.2009.684.
14. Fuccio L, Hassan C, Frazzoni L, Miglio R, Repici A. Clinical outcomes following stent placement in refractory benign esophageal stricture: A systematic review and meta-analysis. Endoscopy. 2016;48(2):141-8. doi: 10.1055/s-0034-1393331.
15. Garbey M, Salmon R, Fikfak V, Clerc CO. Esophageal stent migration: Testing few hypothesis with a simplified mathematical model. Computers in Biology and Medicine. 2016;79:259-65. doi: 10.1016/j.compbiomed.2016.10.024.
16. Conio M, Repici A, Battaglia G, De Pretis G, Ghezzo L, Bittinger M, et al. A randomized prospective comparison of self-expandable plastic stents and partially covered self-expandable metal stents in the palliation of malignant esophageal dysphagia. The American journal of gastroenterology. 2007;102(12):2667. doi: 10.1111/j.1572-0241.2007.01565.x.
17. Food and drugs, 21 C.F.R. § 878.3610 (2019) [regulation on the Internet]. [cited 2020 April 24]. Available from: https://bit.ly/2WfkZlL.
18. Food and Drug Administration: Guidance for the content of premarket notifications for esophageal and tracheal prostheses. Rockville, MD, US Department of Health and Human Services, Food and Drug Administration 1998. Available from :https://bit.ly/3aHt8VH.
19. Duerig T, Tolomeo D, Wholey M. An overview of superelastic stent design. Minimally invasive therapy & allied technologies. 2000;9(3-4):235-46. doi: 10.1080/13645700009169654.

20. Jedwab MR, Clerc CO. A study of the geometrical and mechanical properties of a selfexpanding metallic stent—theory and experiment. Journal of Applied Biomaterials. 1993;4(1):77-85. doi: 10.1002/jab.770040111.
21. Borghi A, Murphy O, Bahmanyar R, McLeod C. Effect of stent radial force on stress pattern after deployment: A finite element study. Journal of Materials Engineering and Performance. 2014;23(7):2599-605. doi: 10.1007/s11665-014-0913-z.
22. Arokiaraj MC, De Santis G, De Beule M, Palacios IF. Finite element modeling of a novel self-expanding endovascular stent method in treatment of aortic aneurysms. Scientific reports. 2014;4:3630.
23. Mozafari H, Dong P, Zhao S, Bi Y, Han X, Gu L. Migration resistance of esophageal stents: The role of stent design. Computers in Biology and Medicine. 2018;100:43-9. doi: 10.1016/j.compbiomed.2018.06.031.
24. Cianchetti M, Laschi C, Menciassi A, Dario P. Biomedical applications of soft robotics. Nature Reviews Materials. 2018;3(6):143-53. doi: 10.1038/s41578-018-0022-y.
25. Fusco S, Sakar MS, Kennedy S, Peters C, Bottani R, Starsich F, et al. An integrated microrobotic platform for on-demand, targeted therapeutic interventions. Advanced Materials. 2014;26(6):952-7. doi: 10.1002/adma.201304098.
26. Laschi C, Mazzolai B, Cianchetti M. Soft robotics: Technologies and systems pushing the boundaries of robot abilities. Science Robotics. 2016;1(1). doi: 10.1126/scirobotics.aah3690.
27. Majidi C. Soft Robotics: A perspective—Current trends and prospects for the future. Soft Robotics. 2014;1(1):5-11. doi: 10.1089/soro.2013.0001.
28. Sim K, Rao Z, Kim HJ, Thukral A, Shim H, Yu C. Fully rubbery integrated electronics from high effective mobility intrinsically stretchable semiconductors. Science Advances. 2019;5(2). doi: 10.1126/sciadv.aav5749.
29. Payne CJ, Wamala I, Bautista-Salinas D, Saeed M, Van Story D, Thalhofer T, et al. Soft robotic ventricular assist device with septal bracing for therapy of heart failure. Science Robotics. 2017;2(12). doi: 10.1126/scirobotics.aan6736.
30. Awad LN, Bae J, O'Donnell K, De Rossi SMM, Hendron K, Sloot LH, et al. A soft robotic exosuit improves walking in patients after stroke. Science translational medicine. 2017;9(400). doi: 10.1126/scitranslmed.aai9084.
31. Ding Y, Kim M, Kuindersma S, Walsh CJ. Human-in-the-loop optimization of hip assistance with a soft exosuit during walking. Science Robotics. 2018;3(15). doi: 10.1126/scirobotics.aar5438.
32. Kikuchi T, Kobayashi H, Michiwaki Y. Development of swallowing robot reproducing hyoid bone and epiglottis during swallowing. Dysphagia. 2009;24(4):467-8.
33. Kou W, Pandolfino J, Kahrilas P, Patankar N. Simulation studies of the role of esophageal mucosa in bolus transport. Biomechanics and Modeling in Mechanobiology. 2017;16(3):1001-9. doi: 10.1007/s10237-016-0867-1.
34. Dirven S, Feijiao C, Weiliang X, Bronlund JE, Allen J, Cheng LK. Design and characterization of a Peristaltic Actuator inspired by esophageal swallowing. IEEE/ASME Transactions on Mechatronics. 2014;19(4):1234-42. doi: 10.1109/TMECH.2013.2276406.
35. Clouse R, Alrakawi A, Staiano A. Intersubject and interswallow variability in topography of esophageal motility. Digestive Diseases and Sciences. 1998;43(9):1978-85. doi: 10.1023/A:1018838710214.
36. Bourne MC. Calibration of rheological techniques used for foods. Journal of Food Engineering. 1992;16(1):151-63. doi: 10.1016/B978-1-85166-877-9.50014-6.

37. Dirven S, Weiliang X, Cheng LK, Allen J. Biomimetic investigation of intrabolus pressure signatures by a Peristaltic Swallowing Robot. IEEE Transactions on Instrumentation and Measurement. 2015;64(4):967-74. doi: 10.1109/TIM.2014.2360800.
38. Chen F-J, Dirven S, Xu W, Li X-N. Soft actuator mimicking human esophageal peristalsis for a swallowing robot. IEEE/ASME Transactions on Mechatronics. 2014;19(4):1300-8. doi: 10.1109/TMECH.2013.2280119.
39. Bhattacharya D, Cheng LK, Xu W. Sparse machine learning discovery of dynamic differential equation of an Esophageal Swallowing Robot. IEEE Transactions on Industrial Electronics. 2020: 67(6): 4711-22. doi: 10.1109/TIE.2019.2928239.
40. Azaouzi M, Makradi A, Belouettar S. Deployment of a self-expanding stent inside an artery: A finite element analysis. Materials & Design. 2012;41:410-20. doi: 10.1016/j.matdes.2012.05.019.
41. Homann N, Noftz MR, Klingenberg-Noftz RD, Ludwig DJDD, Sciences. Delayed complications after placement of self-expanding stents in malignant esophageal obstruction: Treatment strategies and survival rate. Digestive diseases and sciences. 2008;53(2):334-40. doi: 10.1007/s10620-007-9862-9.
42. Homs MYV, Steyerberg EW, Kuipers EJ, van der Gaast A, Haringsma J, van Blankenstein M, et al. Causes and treatment of recurrent dysphagia after self-expanding metal stent placement for palliation of esophageal carcinoma. Endoscopy. 2004;36(10):880-6. doi: 10.1055/s-2004-825855.
43. Ren J, Massey BT, Dodds WJ, Kern MK, Brasseur JG, Shaker R, et al. Determinants of intrabolus pressure during esophageal peristaltic bolus transport. American Journal of Physiology - Gastrointestinal and Liver Physiology. 1993;264(3):G407-G13.doi: 10.1152/ajpgi.1993.264.3.G407.

**Supplementary Materials**

Fig. S1. Schematic of the setup used for measuring stent RF and migration.
Fig. S2. Plot of the stent axial strain as a function of applied chamber pressure in RoSE.
Movie S1. Symmetric contraction of each layer in RoSE.
Movie S2. Peristaltic bolus transport in stent implanted RoSE.
Movie S3. Demonstration of peristaltic wave in quarter version of RoSE.

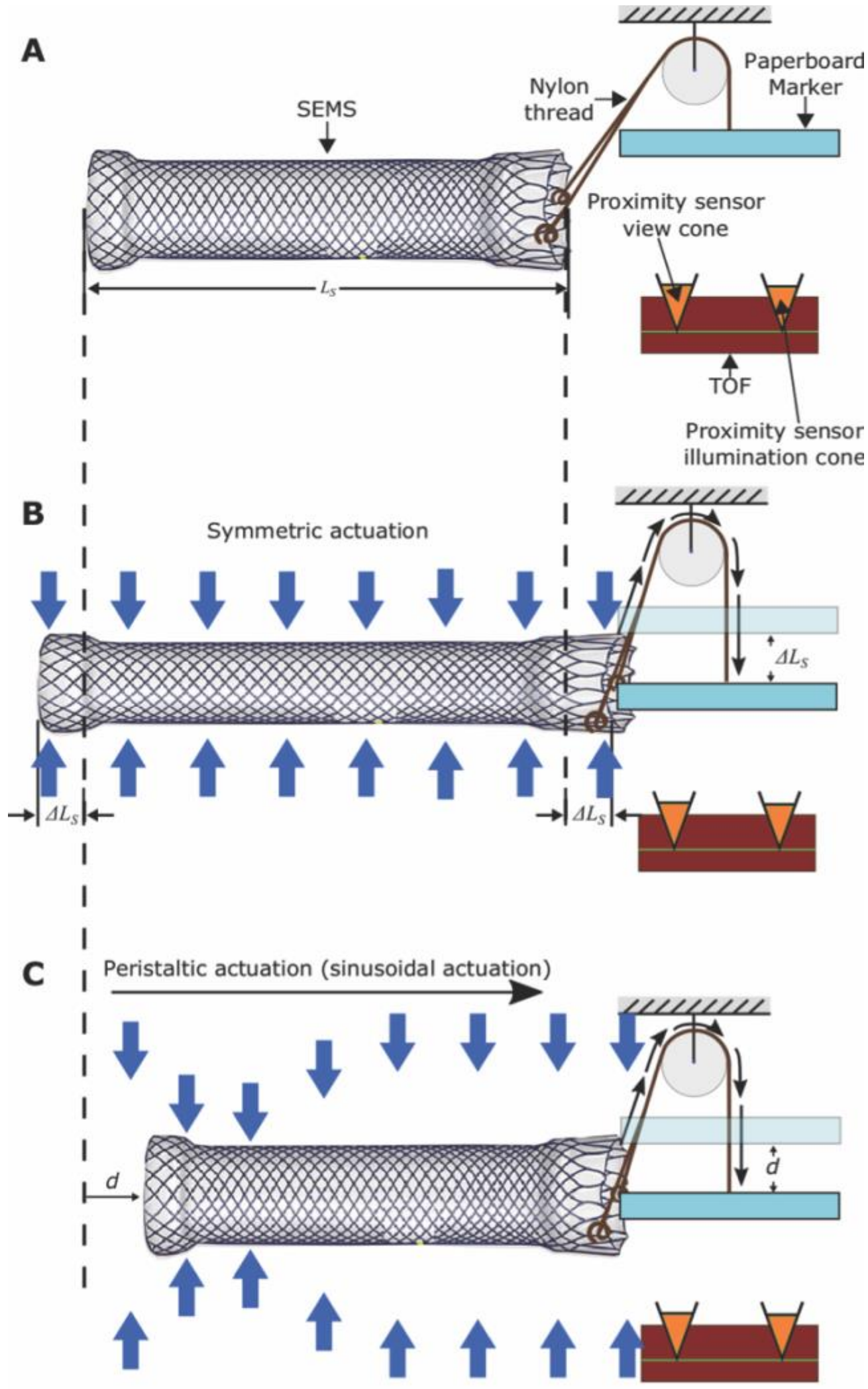

**Fig. S1. Schematic of the setup used for measuring stent RF and migration.** (**A**) A paperboard marker was attached to the distal end of the stent through a pulley. (**B**) Under symmetric contraction of the air chambers layers of the RoSE, the stent elongated axially by $\Delta L_S$ towards its both ends. (**C**) Due to the peristalsis (thick blue arrows), the stent moved by a displacement *d* in the direction of the peristalsis. Along with the stent, the marker also covered a displacement *d* vertically downwards. A time of flight (TOF) sensor was used to record the vertical displacement of the marker.

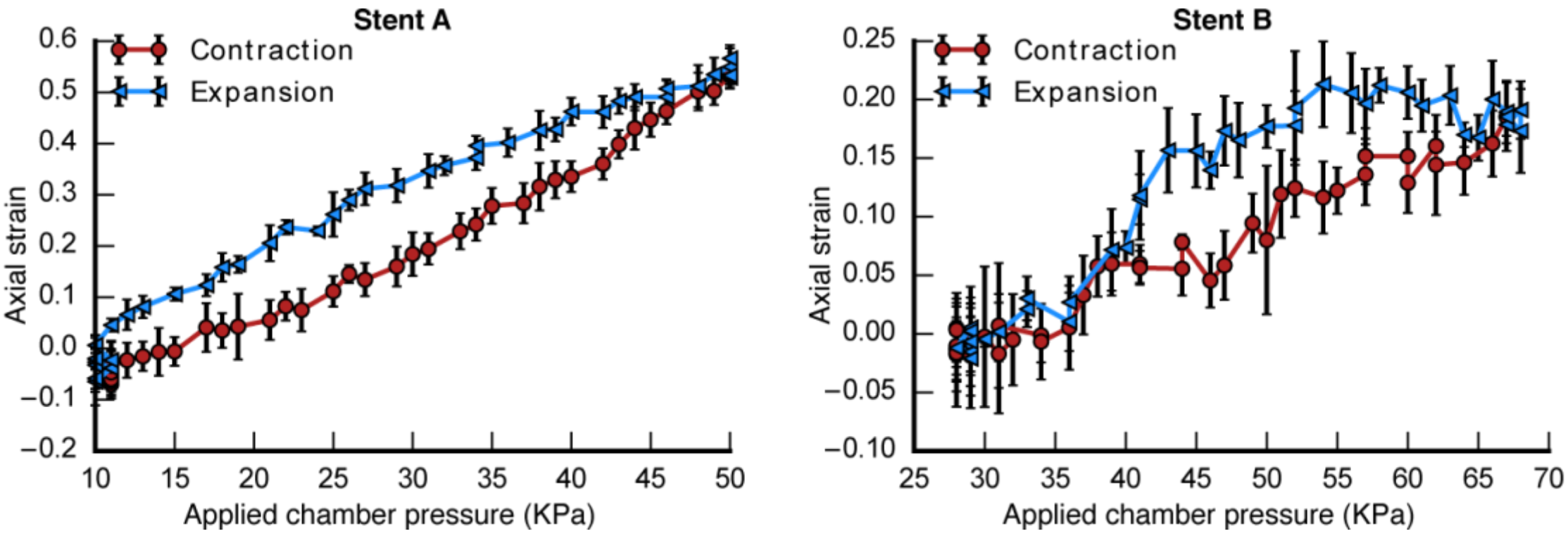

**Fig. S2. Plot of the stent axial strain as a function of applied chamber pressure in RoSE.** With the TOF (time of flight) sensor, the change in length of the stents were measured. The axial (elongation) strain was calculated by the change in stent length per unit initial length (110 mm). For a chamber pressure of 50 ± 0.35 KPa, stent A and B exhibited a 56.60 ± 2 and 17.70 ± 2 % increase in their respective lengths.